\documentclass[4paper]{article}
\usepackage[utf8]{inputenc}
\usepackage{times}
\usepackage{epsfig}
\usepackage{graphicx}
\usepackage{amsmath}
\usepackage{amssymb}
\usepackage{dsfont}
\usepackage{subcaption}
\usepackage{authblk}
\usepackage[square, numbers]{natbib}
\usepackage[a4paper,top=3cm,bottom=2cm,left=3cm,right=3cm,marginparwidth=1.75cm]{geometry}

\begin{document}


\title{Beyond One-hot Encoding: lower dimensional target embedding}

\author[1]{Pau~Rodríguez\thanks{pau.rodriguez@cvc.uab.cat}}
\author[2]{Miguel~A. Bautista\thanks{miguel.bautista@iwr.uni-heidelberg.de}}
\author[3]{Jordi~Gonzàlez\thanks{jordi.gonzalez@cvc.uab.cat}}
\author[1,3]{Sergio~Escalera\thanks{sescalera@cvc.uab.cat}}
\affil[1]{Computer Vision Center, Universitat Autònoma de Barcelona, Spain}
\affil[2]{Heidelberg Collaboratory for Image Processing, Heidelberg University, Germany}
\affil[3]{University of Barcelona, Barcelona, Spain}

\renewcommand\Authands{ and }
\maketitle

\begin{abstract}
Target encoding plays a central role when learning Convolutional Neural Networks. In this realm, One-hot encoding is the most prevalent strategy due to its simplicity. However, this so widespread encoding schema assumes a flat label space, thus ignoring rich relationships existing among labels that can be exploited during training. In large-scale datasets, data does not span the full label space, but instead lies in a low-dimensional output manifold. Following this observation, we embed the targets into a low-dimensional space, drastically improving convergence speed while preserving accuracy. Our contribution is two fold: \emph{(i)} We show that random projections of the label space are a valid tool to find such lower dimensional embeddings, boosting dramatically convergence rates at zero computational cost; and \emph{(ii)} we propose a normalized eigenrepresentation of the class manifold that encodes the targets with minimal information loss, improving the accuracy of random projections encoding while enjoying the same convergence rates. Experiments on CIFAR-100, CUB200-2011, Imagenet, and MIT Places demonstrate that the proposed approach drastically improves convergence speed while reaching very competitive accuracy rates.
\end{abstract}


\section{Introduction}
Convolutional Neural Networks lie at the core of the latest breakthroughs in large-scale image recognition~\cite{ILSVRC15, lin2014microsoft}, at present even surpassing human performance \cite{he2015delving}, applied to the classification of objects \cite{Everingham15}, faces \cite{liu2015faceattributes}, or scenes \cite{zhou2014learning}. Due to its effectiveness and simplicity, one-hot encoding is still the most prevalent procedure for addressing such multi-class classification tasks: in essence, a function $f: \mathbb{R}^p \rightarrow \mathbb{Z}_2^n$ is modeled, that maps image samples to a probability distribution over a discrete set of the $n$ labels of target categories.

Unfortunately, when the output space grows, class labels do not properly span the full label space, mainly due to existing label cross-correlations. Consequently, one-hot encoding might result inadequate for fine-grained classification tasks, since the projection of the outputs into a higher dimensional (orthogonal) space dramatically increases the parameter space of computed models. In addition, for datasets with a large number of labels, the ratio of samples per label is typically reduced. This constitutes an additional challenge for training CNN models in large output spaces, and the reason of slow convergence rates \cite{vijayanarasimhan2014deep}.

In order to address the aforementioned limitations, output embeddings have been proposed as an alternative to the one-hot encoding for training in large output spaces \cite{bengio2010label}: depending on the specific classification task at hand, using different output embeddings captures different aspects of the structure of the output space. Indeed, since embeddings use weight sharing during training for finding simpler (and more natural) partitions of classes, the latent relationships between categories are included in the modeling process. 

According to Akata \emph{et al.} \cite{akata2016label}, output embeddings can be categorized as: 

\begin{itemize}

\item Data-independent embeddings, such as drawing rows or columns from a Hadamard matrix \cite{hsu2009multi}: data-independent embeddings produce strong baselines \cite{frome2013devise}, since embedded classes are equidistant due to the lack of prior knowledge; 

\item Embeddings based on a priori information, like attributes \cite{akata2013label}, or hierarchies \cite{tsochantaridis2005large}: unfortunately, learning from attributes requires expert knowledge or extra labeling effort and hierarchies require a prior understanding of a taxonomy of classes, and in addition, approaches that use textual data as prior do not guarantee visual similarity \cite{frome2013devise}; and 

\item Learned embeddings, for capturing the semantic structure of word sequences (i.e. annotations) and images jointly \cite{weston2010large}. The main drawbacks of learning output embeddings are the need of a high amount of data, and a slow training performance.

\end{itemize}

Thus, in cases where there exist high quality attributes, methods with prior information are preferred, while in cases of a known equidistant label space, data-independent embeddings are a more suitable alternative. Unfortunately, the architectural design of a model is bound to the particular choice among the above-mentioned embeddings. Thus, once a model is chosen and trained using an specific output embedding, it is hard to reuse it for another tasks requiring a different type of embedding.  

In this paper, Error-Correcting Output Codes (ECOC) are proven to be a better alternative to one-hot encoding for image recognition, since ECOCs are a generalization of the three embedding categories~\cite{escalera10}, so a change in the ECOC matrix will not constitute a change in the chosen architecture. In addition, ECOCs naturally enable error-correction, low dimensional embedding spaces \cite{bautista2012minimal}, and bias and variance error reduction~\cite{kong1995error}. 

\begin{figure}[!ht]
\centering
\includegraphics[width=\linewidth]{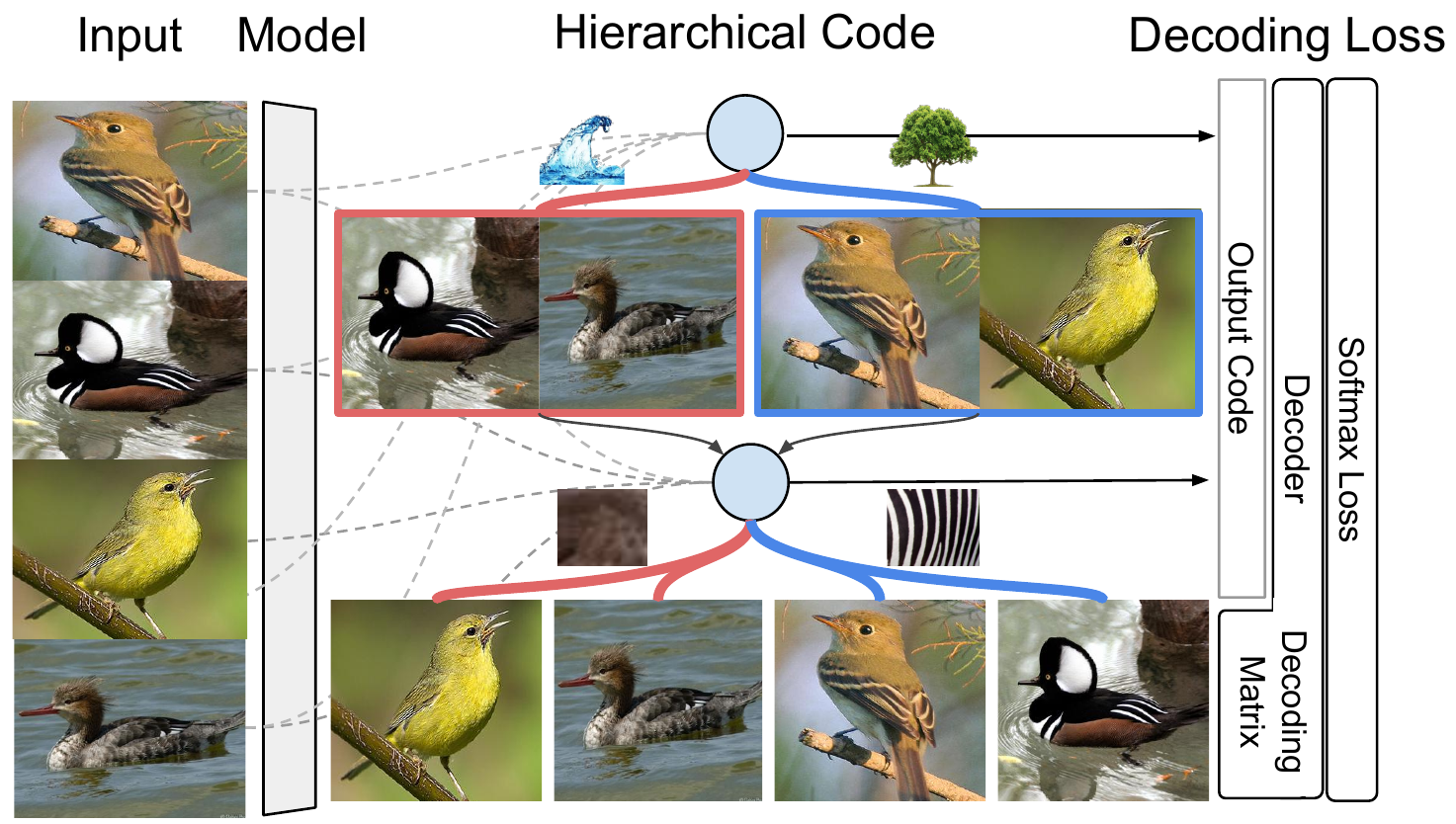}
\caption{This paper proposes to replace the traditional one-hot output scheme of CNNs with a reduced scheme with at least $log_2(k)$ outputs. In addition, when using a hierarchical representation of the data labels, outputs show that the most discriminative attributes to split the target classes have been learned. In essence, a decoder computes the similarities of the "predicted code" in a "code-matrix", and subsequently the output label is then obtained through a softmax layer. The internal code representation is depicted in a tree structure, where each bit of the code corresponds to the actual learned partition from the data, from lower partition cost (aquatic) to higher (stripped).}
\label{fig:model_overview}\vspace{-0.4cm}
\end{figure}

Inspired by the latest advances on ECOCs, we circumvent one-hot encoding by integrating the Error-Correcting Output Codes into CNNs, as a generalization of output embedding. As a result, a best-of-both-worlds approach is indeed proposed: compact outputs, data-based hierarchies, and error correction. Using our approach, training models in low-dimensional spaces  drastically improves convergence speed in comparison to one-hot encoding. Figure \ref{fig:model_overview} shows an overview of the proposed model. 

The rest of the paper is organized as follows: Section \ref{sec:related_work} reviews the existing work most closely related to this paper. Section \ref{sec:method} presents the contribution of the proposed embedding technique, which is two fold: \emph{(i)} we show that random projections of the label space are suitable for finding useful lower dimensional embeddings, while boosting dramatically convergence rates at zero computational cost; and \emph{(ii)} In order to generate partitions of the label space that are more discriminative than the random encoding (which generates random partitions of the label space), we also propose a normalized eigenrepresentation of the class manifold to encode the targets with minimal information loss, thus improving the accuracy of random projections encoding while enjoying the same convergence rates. Subsequently, the experimental results on CIFAR-100 \cite{krizhevsky2009learning},  CUB200-2011 \cite{weinberger2009large}, MIT Places \cite{zhou2014learning}, and ImageNet \cite{ILSVRC15} presented in Section \ref{sec:experiments} show that our approach drastically improves convergence speed while maintaining a competitive accuracy. Lastly, Section \ref{sec:conclusion} concludes the paper discussing how, when gradient sparsity on the output neurons is highly reduced, more robust gradient estimates and better representations can be found.

\section{Related work}\label{sec:related_work}
This section reviews those works on output embeddings most related to ours, in particular those using ECOC.
\medskip
\\
\noindent \textbf{Output Embeddings} Most of the related literature addresses the challenge of zero-shot learning, i.e. training a classifier in the absence of labels. Often, the proposed approaches take into account the attributes of objects \cite{yu2010attribute, rohrbach2011evaluating,kankuekul2012online,akata2016label} related to the different classes through well-known, shared object features.

Due to their computing efficiency based on a divide-and-conquer strategy, output embeddings have been also proven useful for those multi-class classification problems in which testing all possible class labels and hierarchical structures is not feasible  \cite{amit2007uncovering,weinberger2009large,weston2010large,bengio2010label}. Given a large output space, most labels are usually considered instances of a superior category e.g., sunflower and violet are flower plants. In this sense, the inherent hierarchical structure of the data makes divide-and-conquer hierarchical output spaces a suitable alternative to the traditionally flat 1-of-N classifiers. Likewise in the context of language processing, Mikolov \emph{et al.} combine Huffman binary codes and hierarchical soft-max in order to map the most frequent codes to shorter paths in a tree \cite{mikolov2013efficient}. 

Because output embeddings enforce weight sharing, they have been also used when the number of classes is rather large, with no clear inter-class boundaries, and a decaying ratio of the number of examples per class. In this context, in order to reduce the output space, Weston \emph{et al.} proposed WSABIE, an online learning-to-rank algorithm to find an embedding for the labels based on images \cite{weston2011wsabie}.

In the field of large-scale recognition, hierarchical approaches such as using tree-based priors \cite{srivastava2013discriminative}, label relational graphs \cite{deng2014large}, CNN hierarchies \cite{xiao2014error}, and HD-CNNs \cite{yanhd} have been proposed. For example in \cite{lin2015deep} binary hash codes are used for fast image retrieval. However, such hierarchical approaches need to be learned, and cannot be easily interchanged with other embeddings. In addition, for approaches learning codes as latent variables, to find the optimal ones in terms of class separability or error correction is not guaranteed \cite{lin2015deep}. Due to all this, ECOC constitute a better alternative for seamless integration with CNNs, as detailed next.
\medskip
\\
\noindent \textbf{Error-Correcting Output Codes\protect\footnote{We use the standard notation in ECOCs: bold capital letters denote matrices (e.g. $\mathbf{X}$) and bold lower-case letters represent vectors (e.g., $\mathbf{z}$). All non-bold letters denote scalar variables.}} ECOC have been applied in multiple fields such as medical imaging \cite{bai2016learning}, face and facial-feature recognition \cite{windeatt2003boosted,smith2015facial}, and segmentation of human limbs cite{sanchez2015hupba8k+}. ECOCs are a generic divide-and-conquer framework that combines binary partitions to achieve multi-class recognition \cite{dietterich1995solving}. Their core property is the capability to correct errors of binary classifiers using redundancy, while reducing the bias and variance of the ensemble \cite{kong1995error}. Advanced approaches propose to use them as intermediate representations \cite{jiang2016learning}. 

ECOC consist of two main steps: \textit{coding} and \textit{decoding}. The \textit{coding} step consists in assigning a codeword of arbitrary length $k$ to each of the $n$ classes. Codewords are organized in a "code matrix" $\mathbf{M}_{k,n} \in \{-1,1\}$, where each column is a binary partition on the label space in meta-classes. Since there are many possible bi-partitions, the design of the code is central for obtaining discriminative ones. Indeed there are several approaches for generating ECOCs: Exhaustive codes \cite{dietterich1995solving}, BCH codes \cite{bose1960class}, random codes \cite{allwein2000reducing}, and circular ECOC \cite{ghaderi2000circular} are few examples of methods that generate codes independently from the inherent structure of the data.

Although ECOCs can be data-independent and even randomly generated, they can also be learnt from data: Pujol \emph{et al.} propose a discriminant ECOC approach based on hierarchical partitions of the output space \cite{pujol2006discriminant}. Subsequently, Escalera \emph{et al.} \cite{escalera2008subclass} proposed to split complex problems into easier subclasses, embedded as binary dichotomizers in the ECOC framework, easier to optimize. In \cite{crammer2002learnability}, it is also shown Optimal continuous ECOCs can be found by gradient descent. Griffin \& Perona \cite{griffin2008learning} use trees to efficiently handle multi-class problems, which posteriorly Zhang \emph{et al.} improved by finding optimal partitions with spectral ECOCs \cite{zhang2009spectral}.

In the decoding step, a sample $x$ can be decoded as the output of $k$ binary classifiers $\{f_1(x),f_2(x), ..., f_k(x)\}$. Given the predicted code, the class label $y$ corresponds to the closest row in $M_{k,n}$. The most common decoding methods are the Hamming and Euclidean distances but there are more sophisticated approaches such as probabilistic-based decoding, especially with ternary codes~\cite{escalera10}.

Inspired from latest ECOC advances, we propose to integrate output codes in large-scale deep learning problems. In this context, few approaches in the literature have been presented: in \cite{deng2010applying, deng2014large}, CNNs are also used to directly predict the code bits for Optical Character recognition (OCR). We go a step further by: \emph{(i)} showing that the convergence speed in large scale settings with millions of images can be dramatically improved; \emph{(ii)} instead of directly predicting the code bits, we integrate the euclidean decoding with the cross-entropy loss, so that the network does not only optimize individual bits independently but also inter-code distances, which results in error-correction. 

Our approach enhances the convergence of CNNs using random codes, i.e. when the inter-class relationships are not considered. We achieve even lower error rates with data-dependent codes, due to using more efficient data partitions. Similarly, Yang \emph{et al.} also used CNNs to integrate data-independent Hadamard Codes with the Euclidean loss \cite{yang2015deep}. But due to the efficiency of data-dependent codes, our encoding proposal is shown more efficient than \cite{yang2015deep}, by halving the required CNN output size, and eliminating the need of training multiple CNNs to predict code chunks.

\section{Low dimensional target embedding}\label{sec:method}
Figure \ref{fig:model_overview} depicts our proposed model inspired by the ECOC framework \cite{dietterich1995solving} and applied for deep supervised learning. Given a set of $n$ classes, an ECOC consists of a set of $k$ binary partitions of the label space (groups of classes) representing each of the $n$ classes in the dataset. The codes are usually arranged in a design matrix $\mathbf{M} \in \{-1,1\}^{n\times k}$.

Let's define the output of the last layer of a neural network as $z^l$, with $l$ the depth of the network. For the sake of clarity the identity non-linearity $\phi(\cdot)$ is used so that $\mathbf{z}^l = \phi(\mathbf{z}^l)$. Thus, given the weights of the previous layer $\mathbf{\Theta}^{(l-1)}$, and the corresponding bias $\mathbf{b}^{(l-1)}$,  $\mathbf{z}^l$ can be computed as $\Theta^{(l-1)} \mathbf{z}^{(l-1)}+\mathbf{b}^{(l-1)}$. 

In our case, we reduce the output dimensionality of a CNN, i.e. the dimensionality of $\mathbf{z}^l$, from $n$ (the number of classes) to $k$, an arbitrary number of partitions. Then, given a design matrix $\mathbf{M}^{n \times k}$, where each row encodes a class label, the predicted class is obtained by finding the distance of the output with each row of the design matrix $\mathbf{D} = \mathbf{M} - \mathds{1}^\top\mathbf{z}^l$, with $\mathds{1}^\top$ a column vector constituted by ones, and obtaining the label with $argmin(\mathbf{D})$. Then, we seamlessly integrate our proposal in the traditional log-likelihood and softmax loss layer. 

\subsection{Embedding output codes in CNNs}
Given a training set  $\{x_i, y_i\}\ i=1:s$, of image-label pairs, CNNs constitute the state-of-the-art at finding good local minima by empirical risk minimization (ERM) using the cross-entropy as the loss function $J$ by means of backpropagation \cite{lecun1995convolutional}:

$$J(X,Y;\Theta)=-\frac{1}{s} \sum_{i=1}^s [y_i log(\hat{y})_i + (1-y_i)log(1-\hat{y}_i)],$$

where $\hat{y}_i = argmax(\mathbf{h}(\mathbf{z}^l(\mathbf{x}_i))) \in \mathbb{R}$ is the predicted label for the $i^{th}$ example and $y_i \in \{0,1\}$ the ground truth label. Since cross-entropy requires probability distributions, the output of the network $\mathbf{z}^l$ is fed to a softmax layer that assigns a probability score to each of the $n$ possible classes:
$$h(\mathbf{z}^l)_j = \frac{e^{z^l_j}}{\sum^N_{i=1} e^{z^l_i}}, j\in\{1,2,...,n\}.$$

The derivative of the loss function $J$ for gradient descent through backpropagation is known to be:
$$\frac{\delta J}{\delta z^l_i} = y_i - \hat{y}_i.$$

The decoder is introduced between the output $\mathbf{z}^l$ of the network and the softmax function $\mathbf{h}(\mathbf{z}^l)$. Concretely, the negative normalized Euclidean distance $\mathbf{D}(\mathbf{z}^l|\mathbf{M})$ between $\mathbf{z}^l$ and the rows in $\mathbf{M}$ is used, so that the output of the softmax represents the probability of the output of the CNN to be decoded as the $i^{th}, i\in\{1,2,3..,n\}$ output word. 

We reformulate the softmax function $\mathbf{h}(\mathbf{z}^l)$ as $\mathbf{h}(\mathbf{D}(\frac{\mathbf{z}^l}{||\mathbf{z}^l||_2}))$, with the variable change of $\mathbf{D}(\frac{\mathbf{z}^l}{||\mathbf{z}^l||_2})$ by $\mathbf{D}(\mathbf{U})$ (with $\mathbf{U}(\mathbf{z})$ the normalized vector). The derivative of the loss can be computed using the chain rule:
$$ \frac{\delta J(\mathbf{D},Y;\Theta)}{\delta \mathbf{z}} = \frac{\delta J(\mathbf{D},Y;\Theta)}{\delta \mathbf{D}}\frac{\delta \mathbf{D}}{\delta \mathbf{U}}^{(1)} \frac{\delta \mathbf{U}}{\delta \mathbf{z}^l}^{(2)}. $$

We now calculate:

\begin{equation}
\frac{\delta \mathbf{D}}{\delta \mathbf{U}} = \frac{\delta }{\delta \mathbf{U}} \frac{-1}{2}(\mathbf{M}-\mathds{1}^\top\mathbf{U})(\mathbf{M}-\mathds{1}^\top\mathbf{U})^\top = \mathbf{M}-\mathds{1}^\top\mathbf{U},
\label{eq:norm_derivative}
\end{equation}

\begin{equation}
\frac{\delta \mathbf{U}}{\delta \mathbf{z}} = \frac{\delta }{\delta \mathbf{z}} \frac{\mathbf{z}}{||\mathbf{z}||_2} = \frac{\mathbf{I}||\mathbf{z}||_2 - \mathbf{z}\mathbf{U}^\top}{||\mathbf{z}||_2}.
\label{eq:distance_derivative}
\end{equation}

Given eq. \ref{eq:norm_derivative} and \ref{eq:distance_derivative}, it is possible to compute the derivative of the cross-entropy with the new decoding loss $\hat{J}$:

\begin{equation}
\frac{\delta \hat{J}}{\delta \mathbf{z}^l} = (\mathbf{Y} - \mathbf{\hat{Y}})[(\mathbf{M}-\mathds{1}^\top\mathbf{U})\frac{\mathbf{I}||\mathbf{z}^l||_2 - \mathbf{z}^l\mathbf{U}^\top}{||\mathbf{z}^l||_2}]^\top.
\end{equation}

Provided the amount of computation that can be shared from the forward pass to the backward pass, this process does not slow-down the training phase. In fact, the cost is compensated by \emph{(i)} the shrinkage of $\mathbf{z}$, which also results in a reduction of the number of network parameters, and \emph{(ii)} the increase of convergence speed. 

The convergence speed increases because reducing the output layer results in parameter sharing, which produces more robust gradient estimates. The explanation is that the softmax function distributes the probabilities among a high number of neurons. Thus, the the gradient $\delta J=y_i - \hat{y}_i$ is zero for most outputs because $y_i=1$ only once in the ground truth vector, and  $\mathbb{E}(\hat{y}_j) = \frac{1}{n}$. Given that the network is certain about the output $i'$, the expected output for the rest of the outputs is even smaller  $\mathbb{E}(\hat{y}_{j\neq i'}) = \frac{1-y_{i'}}{n-1}$. 

In other words, output layers with huge number of outputs and smaller mini-batch size can only update the weights of few output units per iteration, since activation expected value is virtually zero. Thus, the gradients for these outputs are either zero or based on too few examples. This leads to noisy estimates to the real loss surface. As a result, reducing the output space with our method increases the ratio of activations per mini-batch, helping to obtain more robust gradient estimates and increasing convergence speed, reduces the mini-batch size, and thus the memory requirements.

\subsection{Connections with Normalized Cuts}
CNNs trained with our approach are robust and fast even when drawing codes from a normal distribution. The reason is the fact that random gaussian matrices tend to follow the coding properties described in the literature  \cite{dietterich1995solving,hastie1998classification}, such as row and column orthogonality. For most large datasets the label space follows a hierarchical structure and defining random partitions of the label space is rather unnatural. In order to find the most simple partitions we use an eigenrepresentation of the class manifold based on the class similarities found in the dataset. Concretely, solving the normalized cut (Ncut) problem on the class similarity graph is a way of obtaining $n$ uncorrelated low-cost partitions, with $n$ the number of classes \cite{shi2000normalized}. The NCut can be approximated by solving the eigendecomposition of the normalized Laplacian of the class similarity matrix $\mathbf{L_M}$:
$$\mathbf{L_G} = \mathbf{D}^{\frac{1}{2}}(\mathbf{D}-\mathbf{M})\mathbf{D}^{\frac{-1}{2}}=\lambda \mathbf{V},$$

where $\mathbf{M}$ is the class similarity matrix, $\mathbf{D}$ is the degree matrix, $\mathbf{\lambda_i}$ are the eigenvalues in ascending order and $\mathbf{v_i}$, the corresponding eigenvectors $i \in \{0,1,2,...,k\}$. Given that $\lambda_0 = 0$, the eigenvectors $\mathbf{v}_i, i\in \{1,...,k\}$ constitute the partitions ordered by the Ncut cost. As explained in \cite{zhou2014learning}, this kind of codes have desirable properties such as balancing, orthogonality, lower error bounds due to the separability maximization, and similarity preserving, i.e. similar classes have similar codes. We show that training CNNs to predict the embedded target, together with this data-based codes, exhibit lower error rates than using random codes. Contrary to \cite{zhang2009spectral}, we do not threshold the eigenvectors so as to obtain a binary code but we interpret the values as likelihoods.

In the following section, we provide empirical evidence confirming that CNNs trained with our proposed methodology on CIFAR-100, CUB-200, MIT Places, and Imagenet have faster convergence rates (with comparable or better recognition rates), even with smaller mini-batch size, than their one-hot counterparts.

\section{Experiments}\label{sec:experiments}

To validate our approach, we perform a thorough analysis of the advantages of embedding output codes in CNN models over different state-of-the-art datasets. First, we describe the considered datasets, methods and evaluation.

\subsection{Datasets}
We first experiment the ImageNet 2012 Large-Scale
Visual Recognition Challenge (ILSVRC-2012) \cite{ILSVRC15} and the MIT Places-205 \cite{zhou2014learning} datasets. 
ImageNet consists of 1.2M images, and 50K validation images with 10K object classes. MIT Places is constituted by 2.5M images from 205 scene categories for training, and 100 images for category for testing.

Subsequently we experiment on the CIFAR-100 \cite{krizhevsky2009learning} and the Caltech-UCSD Birds-200-2011  \cite{welinder2010caltech}. CIFAR-100 consists of 50K $32\times 32$ images for training, and 10K $32\times 32$ images for testing belonging to 10 coarse categories and 100 fine-grained categories. 
CUB-200 contains 11,788 images (5,994 images for training and 5,794 for test) of 200 bird species, each image annotated with 15 part locations, 312 binary attributes, and 1 Bounding Box.

\subsection{Methods and evaluation}
We use standard state-of-the-art models to evaluate the contribution of the proposed target embedding procedure instead of comparing with state-of-the-art results on the considered datasets. Note that any model, including more recent and powerful state-of-the-art architectures, can benefit from our target embedding methodology.

As a proof of concept, we first validate data-independent codes on the Imagenet and MIT Places datasets. Concretely, we retrain with our approach the \texttt{fc7} and \texttt{fc8} layers of an Alexnet model \cite{krizhevsky2012imagenet} pre-trained on the respective datasets. Concretely, we randomly reinitialize their weights and train them using SGD with a global learning rate (\texttt{lr}) of $0.001$, and the specific \texttt{lr} of the reinitialized layers is multiplied by $10$. 

Then, we demonstrate the advantages of data-dependent codes on the fine-grained CIFAR-100 and CUB-200 2011. For CIFAR-100, we use the \texttt{cifar\_quick} models found in the Caffe framework \cite{jia2014caffe}. The network is initialized with noise sampled from a gaussian distribution, and the model is trained for 100 epochs.
Fine-tuning on CUB-200 is performed with the same pre-trained model of the Imagenet experiments for $30$ epochs, and the \texttt{lr} is divided by $10$ after $15$ epochs. 

Experiments with the standard Alexnet CNN \cite{krizhevsky2012imagenet} (caffe version \cite{jia2014caffe}) on Imagenet, and MIT Places, prove that CNNs trained with random codes and our approach show faster convergence rates than using one-hot encoding, especially for small mini-batch sizes, while matching one-hot in performance for bigger mini-batch sizes. Thus, the proposed data-dependent encoding approach performs better than using random codes for fine-grained datasets, with fuzzy inter-class boundaries, essentially because random codes alone do not take into account the correlation of attributes. 

\begin{figure}[!t]
\centering
\begin{subfigure}[t]{0.49\linewidth}
\centering
\includegraphics[width=\textwidth]{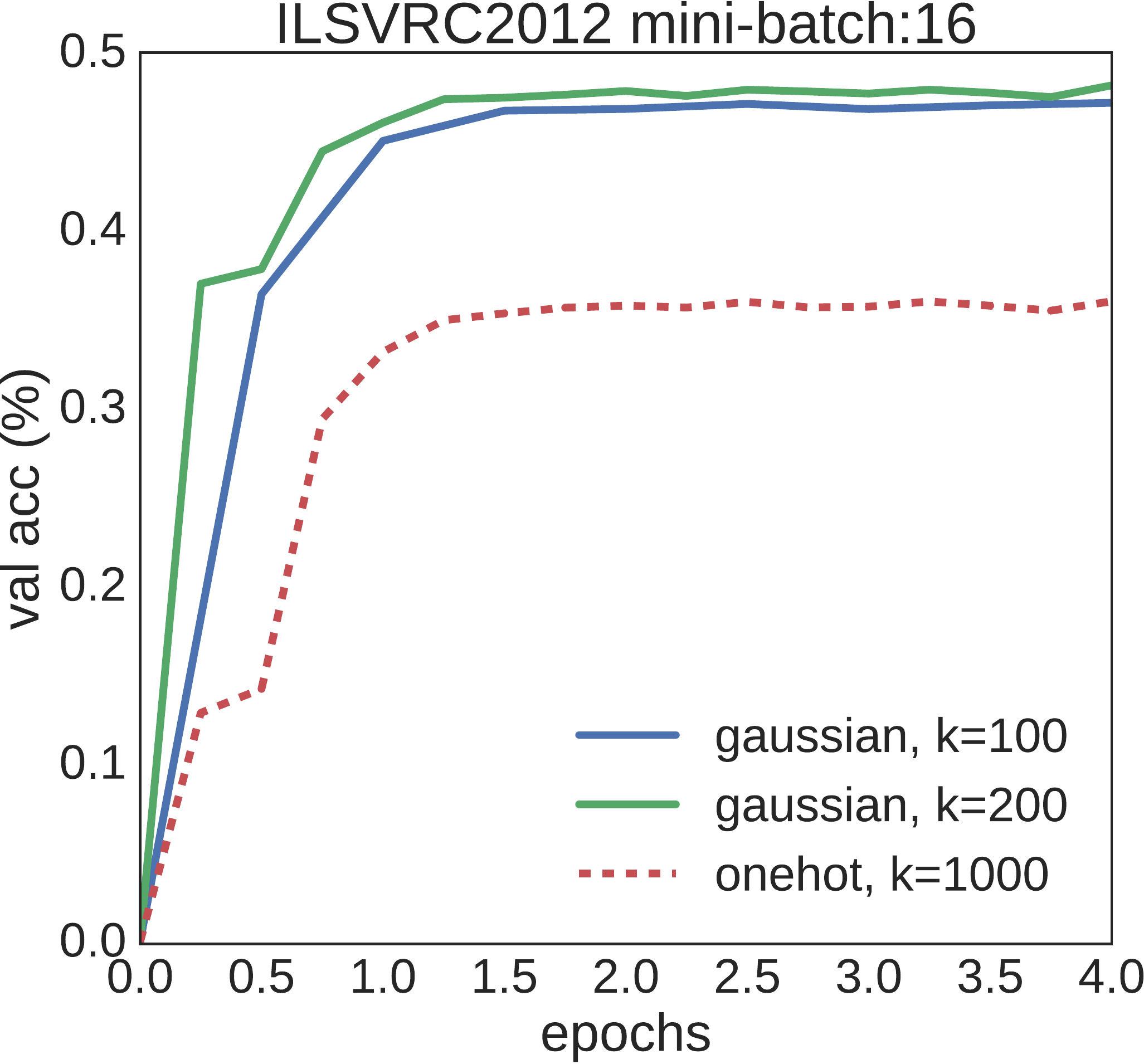}
\caption{ILSVRC2012, \texttt{bs=16}}
\end{subfigure}
\begin{subfigure}[t]{0.49\linewidth}
\centering
\includegraphics[width=\textwidth]{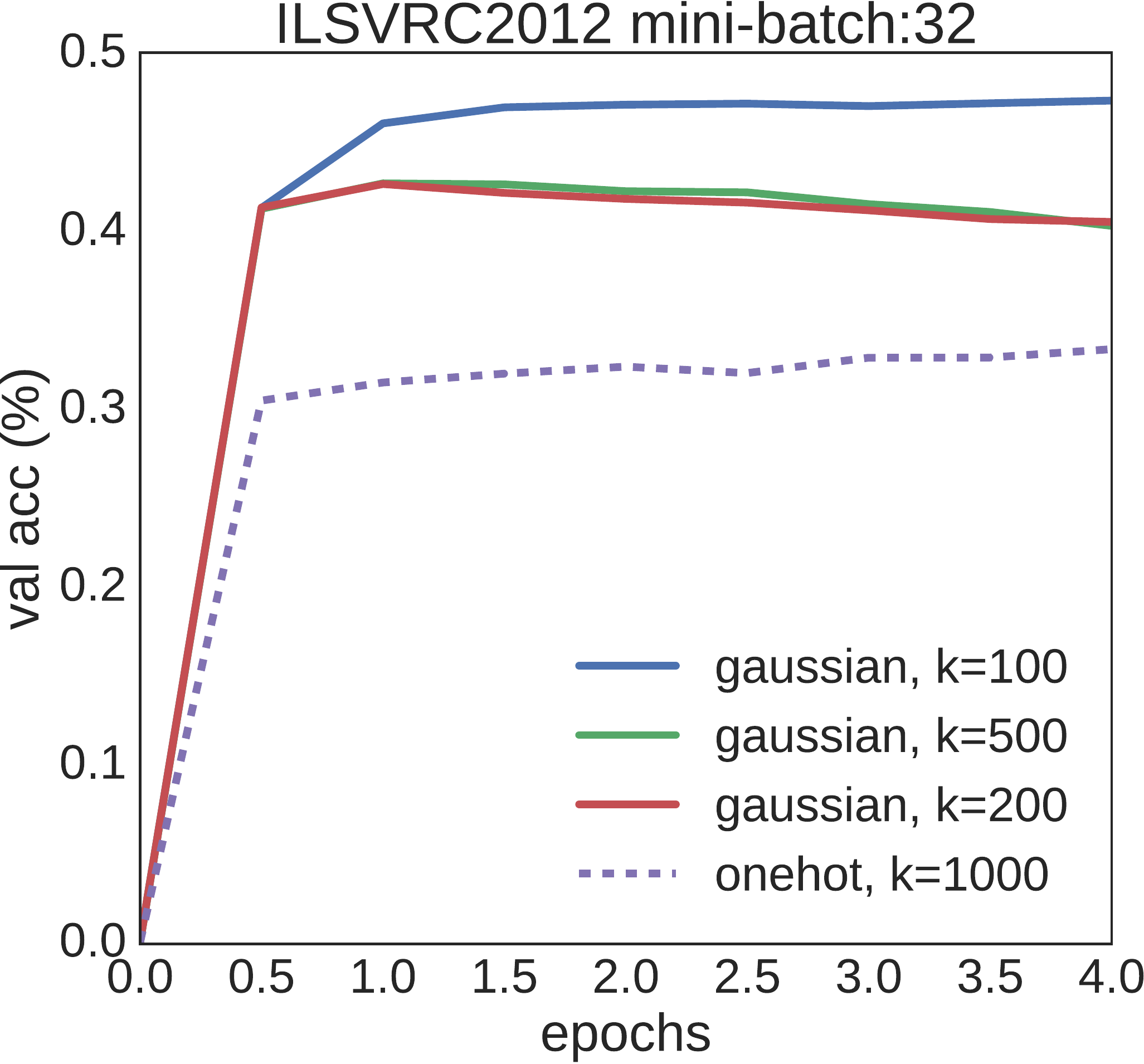}
\caption{ILSVRC2012, \texttt{bs=32}}
\end{subfigure}
\begin{subfigure}[t]{0.49\linewidth}
\centering
\includegraphics[width=\textwidth]{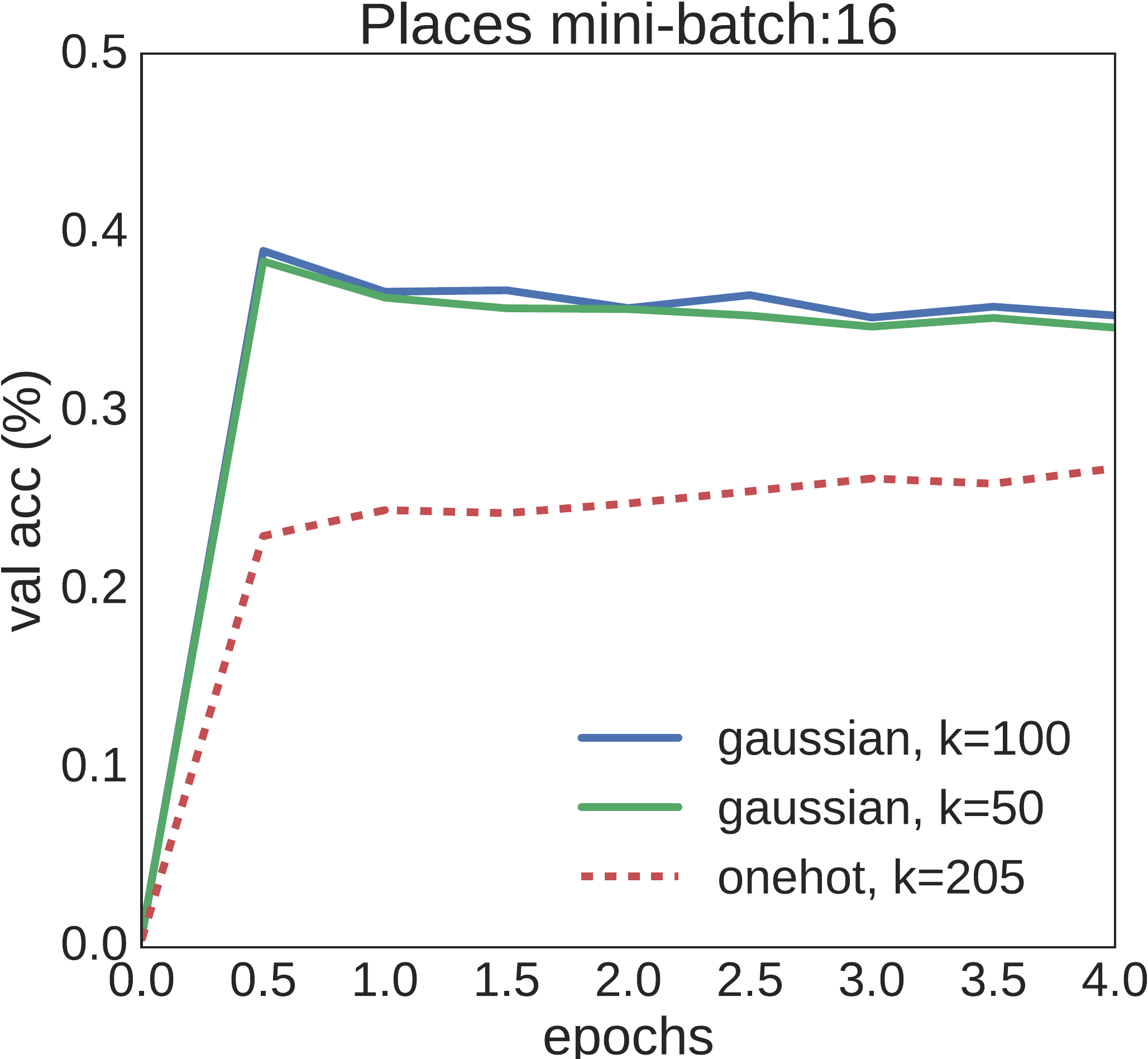}
\caption{MIT Places, \texttt{bs=16}}
\end{subfigure}
\begin{subfigure}[t]{0.49\linewidth}
\centering
\includegraphics[width=\textwidth]{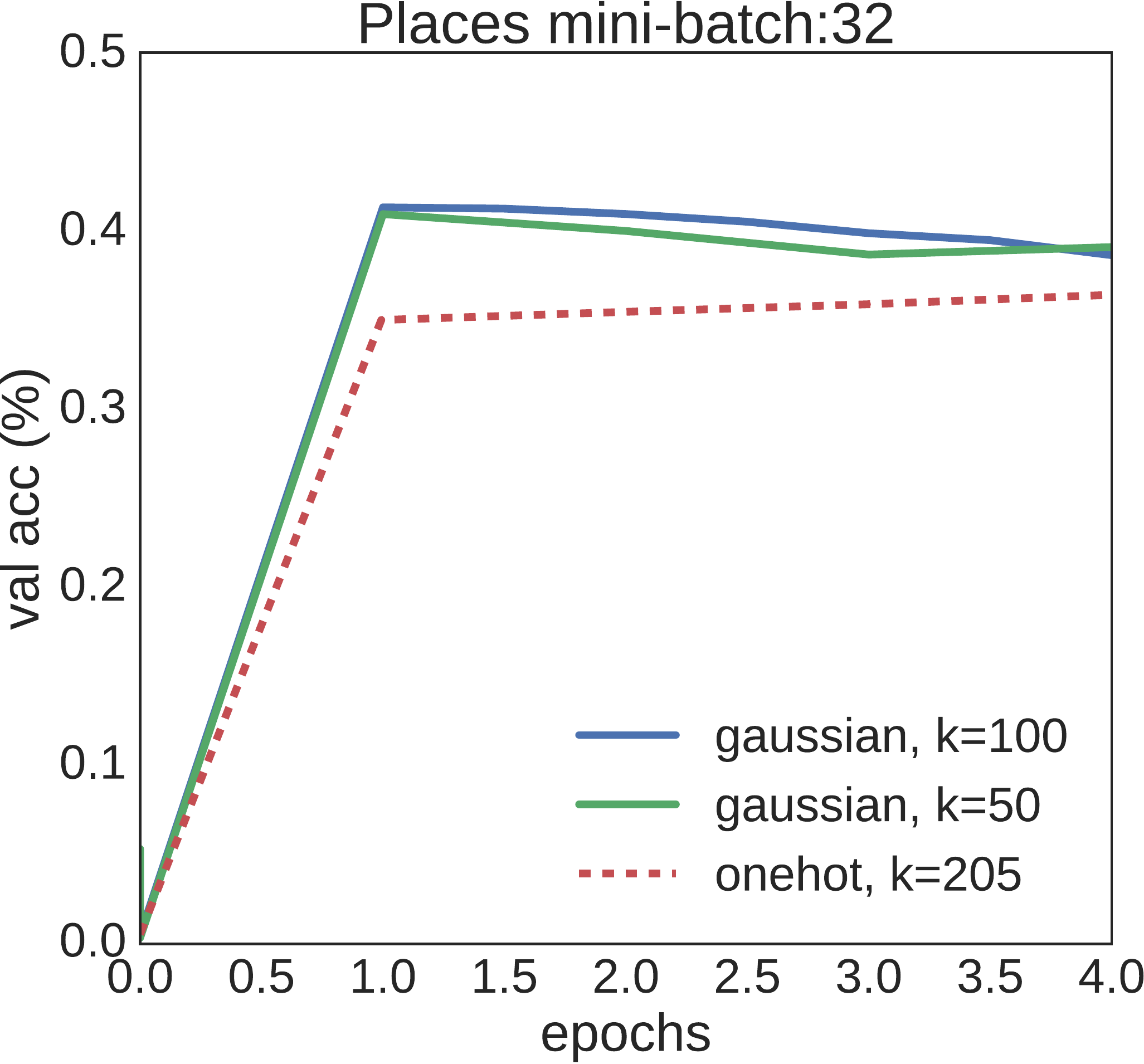}
\caption{MIT Places, \texttt{bs=32}}
\end{subfigure}
\caption{\textbf{Validation accuracy on ILSVRC2012 and MIT Places.} Using output codes randomly sampled from a normal distribution results in faster convergence, especially for small mini-batch sizes (a,c)}
\label{fig:imagenet_bs}\vspace{-0.4cm}
\end{figure}

\subsection{Random codes for faster convergence}

Output encodings allow to embed sparse output spaces into compact representations. For instance, codes generated with the dense random strategy only need $k=10log(n)$ bits \cite{allwein2000reducing} to encode $n$ classes. An inherent property of one-hot encoding is the output activation sparsity for huge output spaces. Given a randomly initialized CNN with one-hot encoding, provided that the output neurons follow a uniform distribution, the probability assigned to each class will be $\frac{1}{n},\ n=\#Classes$, which tends to $0$ for $n \rightarrow \infty$. In the final stages of training, the situation will persist since just an extremely small ratio of the neurons activate, i.e. a small subset of the neurons show high probability for the predicted class while the residual probability mass is spread over a much larger number of neurons. 

Thus, it can be coarsely estimated that the update probability of the parameters associated to an output neuron during an SGD step is related to the ratio $\rho=\frac{bs}{n}$, with mini-batch size $bs$, being $\rho=256\cdot 10^{3}$ for Alexnet trained on Imagenet, provided that $p(Y=n_i)=p(Y=n_j),\ i\neq j$. In other words, given a label, sampling more images increases the probability of that label being in the set of samples, and drawing less samples than the number of labels ensures that at least $n-s$ labels will not be seen during the update.

Figure \ref{fig:imagenet_bs} shows the resulting validation accuracy when training Alexnet on the ILSVRC2012 and MIT Places for different mini-batches and a random code sampled from $\mathcal{N}(0,1)$.
As it can be seen, models trained with our approach converge faster than those trained with one-hot encoding.

\subsection{Using data-based encodings}
In order to adapt to fine-grained settings, i.e. with high inter-class correlations, and few examples per class, we propose to generate the output codes using the eigenvectors of the normalized Laplacian of the class similarity matrix. Since this eigendecomposition generates the most discriminating, hierarchical partitions based on the data, models trained with this data-dependent codes result in higher accuracy bounds than the random counterparts. 

\begin{figure}[!t]
\begin{subfigure}[t]{0.47\linewidth}
\centering
\includegraphics[width=\textwidth,height=0.9\textwidth]{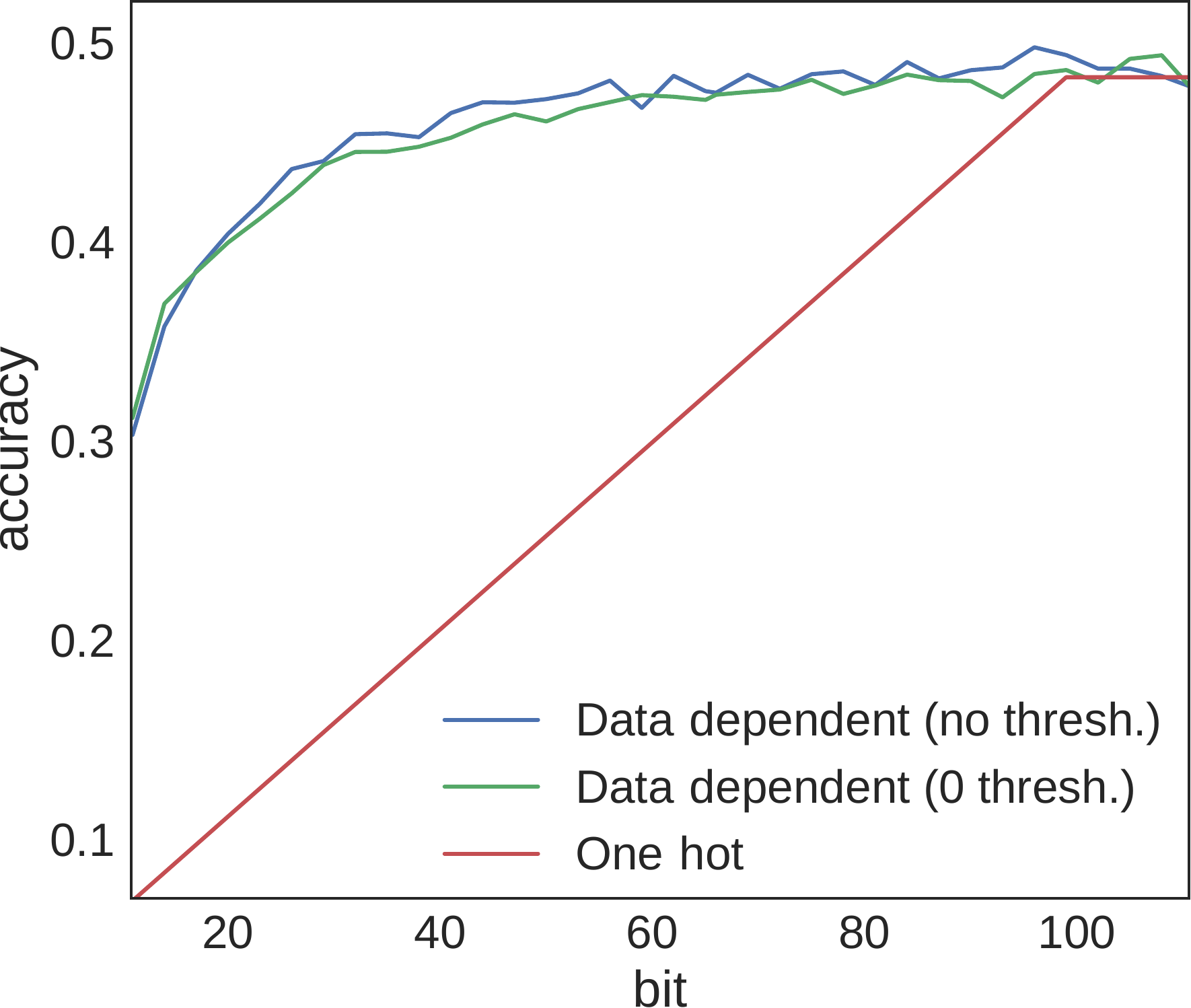}
\caption{CIFAR100}
\label{fig:cifar100_add_bits}
\end{subfigure}
\begin{subfigure}[t]{0.47\linewidth}
\centering
\includegraphics[width=\textwidth,height=0.9\textwidth]{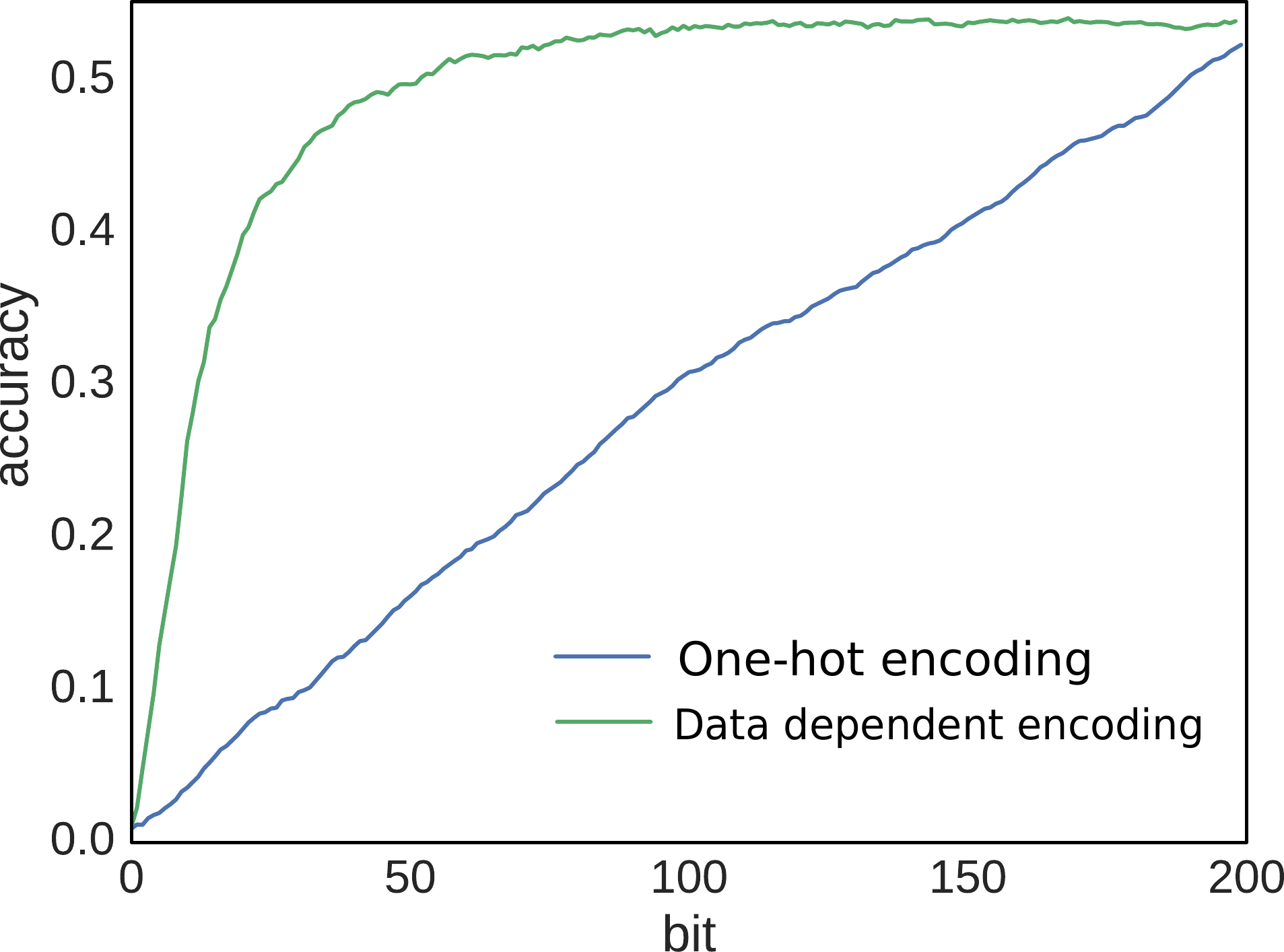}
\caption{CUB200}
\label{fig:cub_add_bits}
\end{subfigure}
\caption{Classification accuracy based on the number of the code bits. As expected, the same amount of information is encoded for each of the one-hot bits while the same results are obtained with just the $25\%$ of the data-based codes.}
\label{fig:add_bits}\vspace{-0.2cm}
\end{figure}

To confirm the aforementioned advantages of using data-dependent codes we choose to experiment on the well-established CIFAR-100 and CUB-200 2011 fine-grained datasets. see Fig. \ref{fig:add_bits}. We use CIFAR-100 for fast experimentation, and then we apply the best setting to CUB-200.

\hfil \\
\noindent \textbf{CIFAR-100}. First, we evaluate different procedures for generating the codes: 
\begin{enumerate}\small
\setlength\itemsep{0.25em}
\item One-hot. A vector of $n-1$ zeros and a one at the target position (with $n$ the number of classes).\vspace{-0.2cm}
\item Dense random \cite{allwein2000reducing}. Sampling the matrix with the most uncorrelated rows and columns from $\mathcal{U}(0,1)$.\vspace{-0.2cm}
\item Gaussian. Sampling matrices from a normal distribution.\vspace{-0.2cm}
\item Data-based. Constructing the code matrix from the eigenvalues of the class similarity Laplacian.
\end{enumerate}

Note that Gaussian and Data-based codes are composed of real numbers and a thresholding function should be applied for obtaining binary partitions. We test thresholding at zero and the median of the rows of the code matrix. Additionally, we test the raw values, interpreting them as the likelihood of the $k^{th}$ metaclass to be present in the $n^{th}$ class. 

\begin{table}[!t]
\centering
\makebox[\textwidth][c]{\begin{tabular}{l|c|c|c|c|c|c|c|c|c|c|c|c|}
\hline
\multicolumn{1}{|l|}{\textbf{Code}} & \multicolumn{3}{c|}{One-hot} & \multicolumn{9}{c|}{Gaussian} \\ \hline
\multicolumn{1}{|l|}{\textbf{Binarization}} & \multicolumn{3}{c|}{-} & \multicolumn{3}{c|}{-} & \multicolumn{3}{c|}{Zero} & \multicolumn{3}{c|}{Median} \\ \hline
\multicolumn{1}{|l|}{\textbf{Length}} & 66 & 100 & 200 & 66 & 100 & 200 & 66 & 100 & 200 & 66 & 100 & 200 \\ \hline
\multicolumn{1}{|l|}{\textbf{Accuracy (\%)}} & 32.4 & 49.2 & - & 44.9 & 44.8 & 44.8 & 45.0 & 47.1 & 49.1 & 45.6 & 47.8 & 48.4 \\ \hline
 & \multicolumn{3}{c|}{Dense Random} & \multicolumn{9}{c|}{Data-dependent} \\ \cline{2-13} 
 & - & - & - & \multicolumn{3}{c|}{-} & \multicolumn{3}{c|}{Zero} & \multicolumn{3}{c|}{Median} \\ \cline{2-13} 
 & 66 & 100 & 200 & 66 & 99 & 200 & 66 & 99 & 200 & 66 & 99 & 200 \\ \cline{2-13} 
 & 43.9 & 44.5 & 44.3 & 48.0 & 50.0 & 49.7 & 46.7 & 49.0 & 48.9 & 47.4 & 47.8 & 49.7 \\ \cline{2-13} 
\end{tabular}}
\caption{\textbf{Influence of code designs on CIFAR-100.} Dense output encodings are more robust than One-hot to the loss of bits. As expected, data-based codes outperform the rest of encodings ($50\%$), especially when no threshold is applied to binarize the code. }
\label{tab:cifar100_ablation}
\end{table}

\begin{figure*}[!t]
\centering
\begin{subfigure}[t]{0.32\textwidth}
\includegraphics[height=4cm,width=\textwidth]{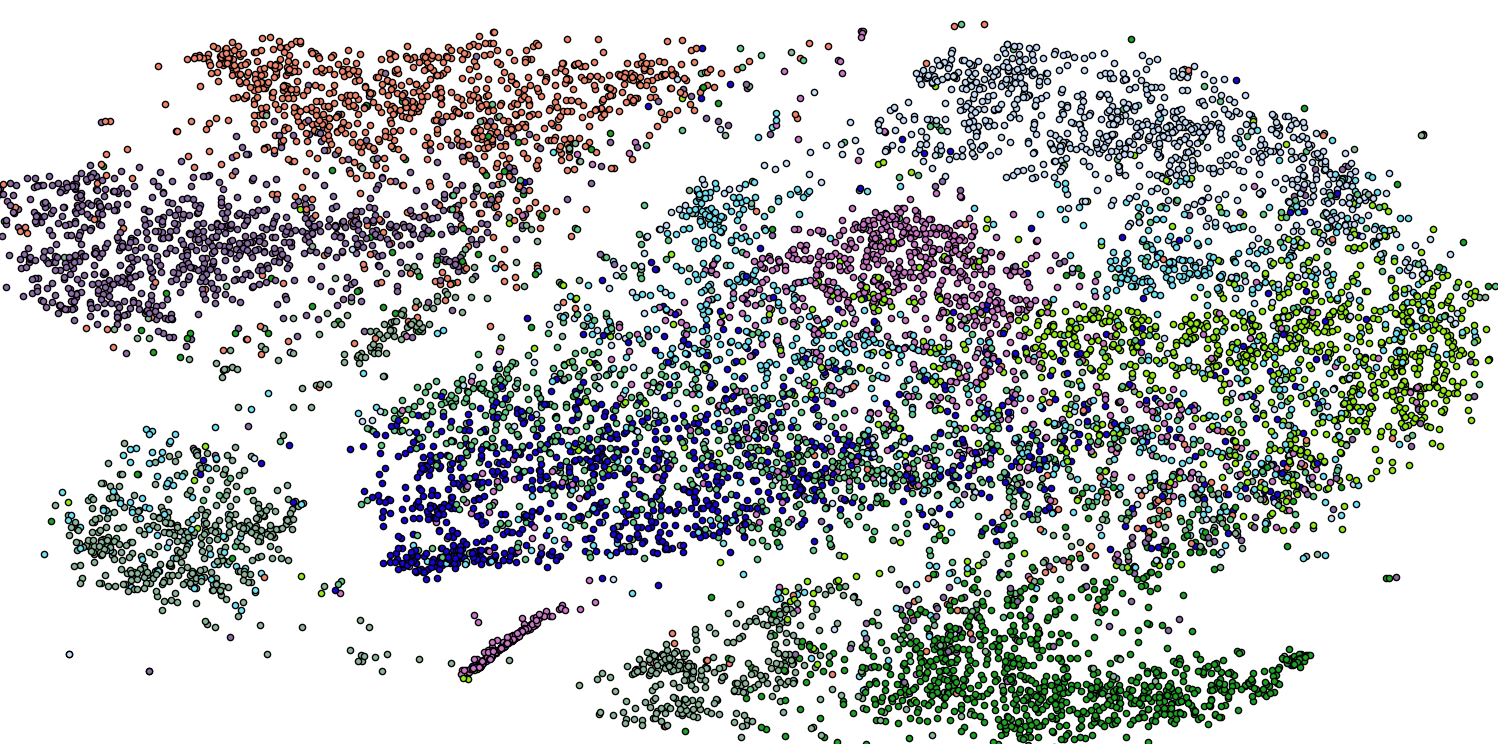}
\caption{One-hot.}
\end{subfigure}
\begin{subfigure}[t]{0.32\textwidth}
\includegraphics[height=4cm,width=\textwidth]{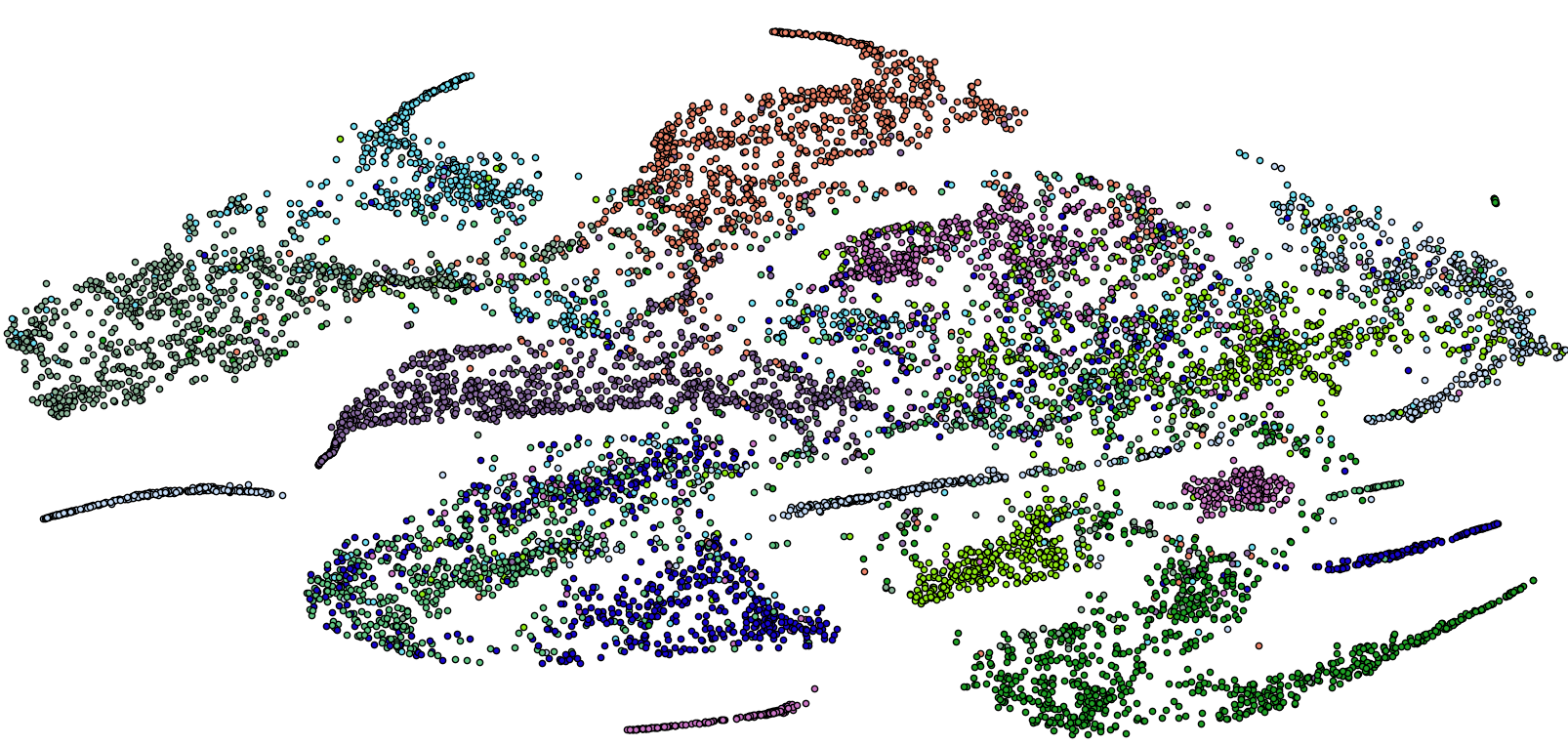}
\caption{Dense random.}
\end{subfigure}
\begin{subfigure}[t]{0.32\textwidth}
\includegraphics[height=4cm,width=\textwidth]{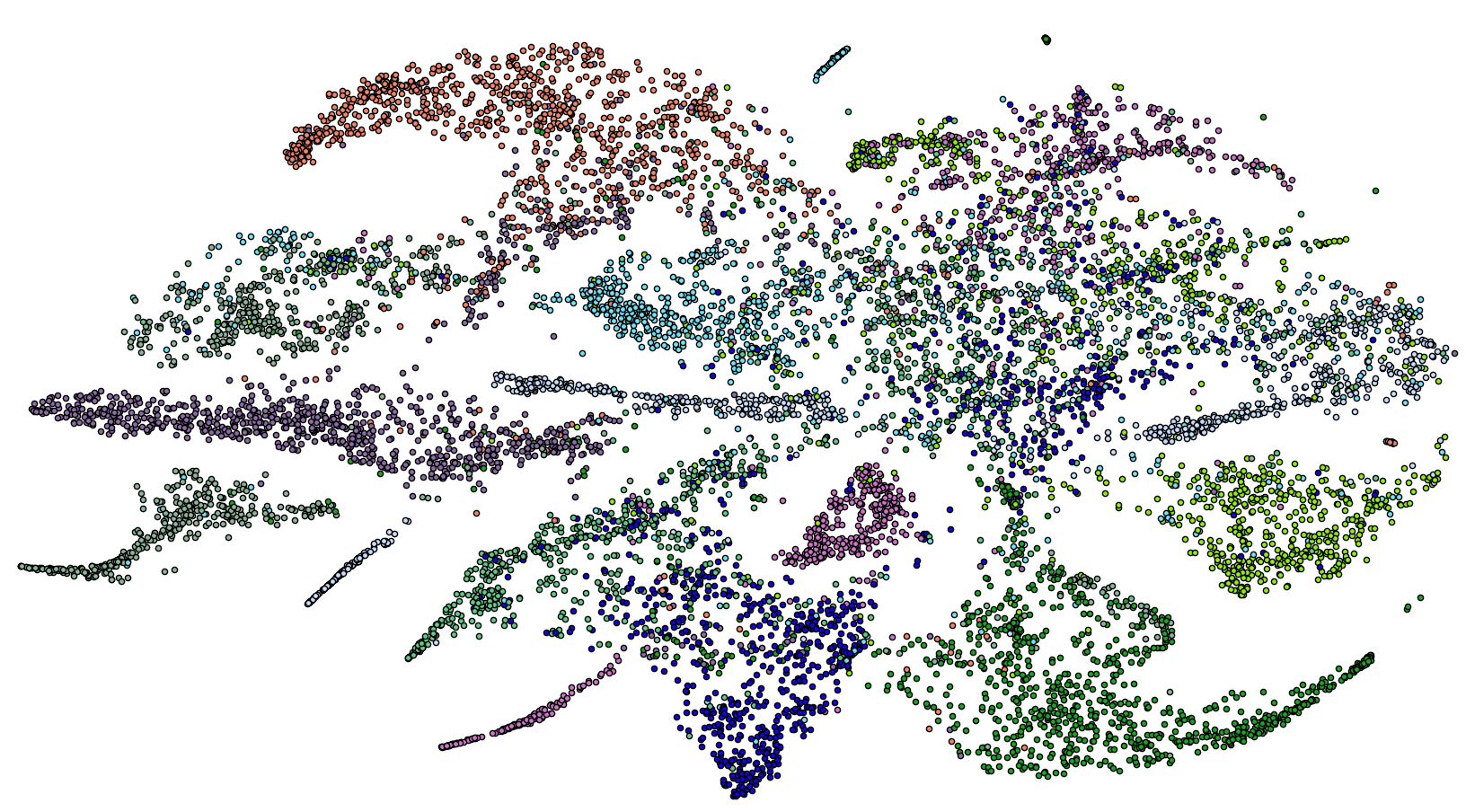}
\caption{Data-based.}
\end{subfigure}
\caption{\textbf{T-sne visualization on CIFAR-100} on the ten coarse categories for the hidden fc layer of a CNN trained with (a) one-hot encoding, (b) an output code generated with the dense random strategy, and (c) a data-based code. }
\label{fig:cifar10_tsne}
\end{figure*}

As it can be seen in table \ref{tab:cifar100_ablation}, output encodings are more robust, losing a smaller percentage of the accuracy when the number of code-bits are halved, while one-hot scales linearly with the number of bits, see \ref{fig:cifar100_add_bits} for a detailed analysis. In addition, data-based codes find the more discriminative partitions, resulting in better accuracy than the rest of the encodings. Moreover, keeping the raw values of the eigenvectors provides additional information about the likelihood of a metaclass to be present in a certain class, resulting in more robust predictions.
Since output codes are based on binary partitions, they constrain the learning so that features are encoded to fall into hyperplanes.

In figure \ref{fig:cifar10_tsne} we show the 2D projection of those hyperplanes using t-sne. Note the higher overlapping of samples from different classes displayed on the target embedding space of 1-hot in comparison to dense and data-dependent alternatives. In particular, the proposed eigendecomposition of the output space shows a more discriminative splitting of the data samples according to their labels.

\begin{figure*}[!t]
\centering
\begin{subfigure}[t]{0.45\textwidth}
\includegraphics[width=\textwidth, height=1\textwidth]{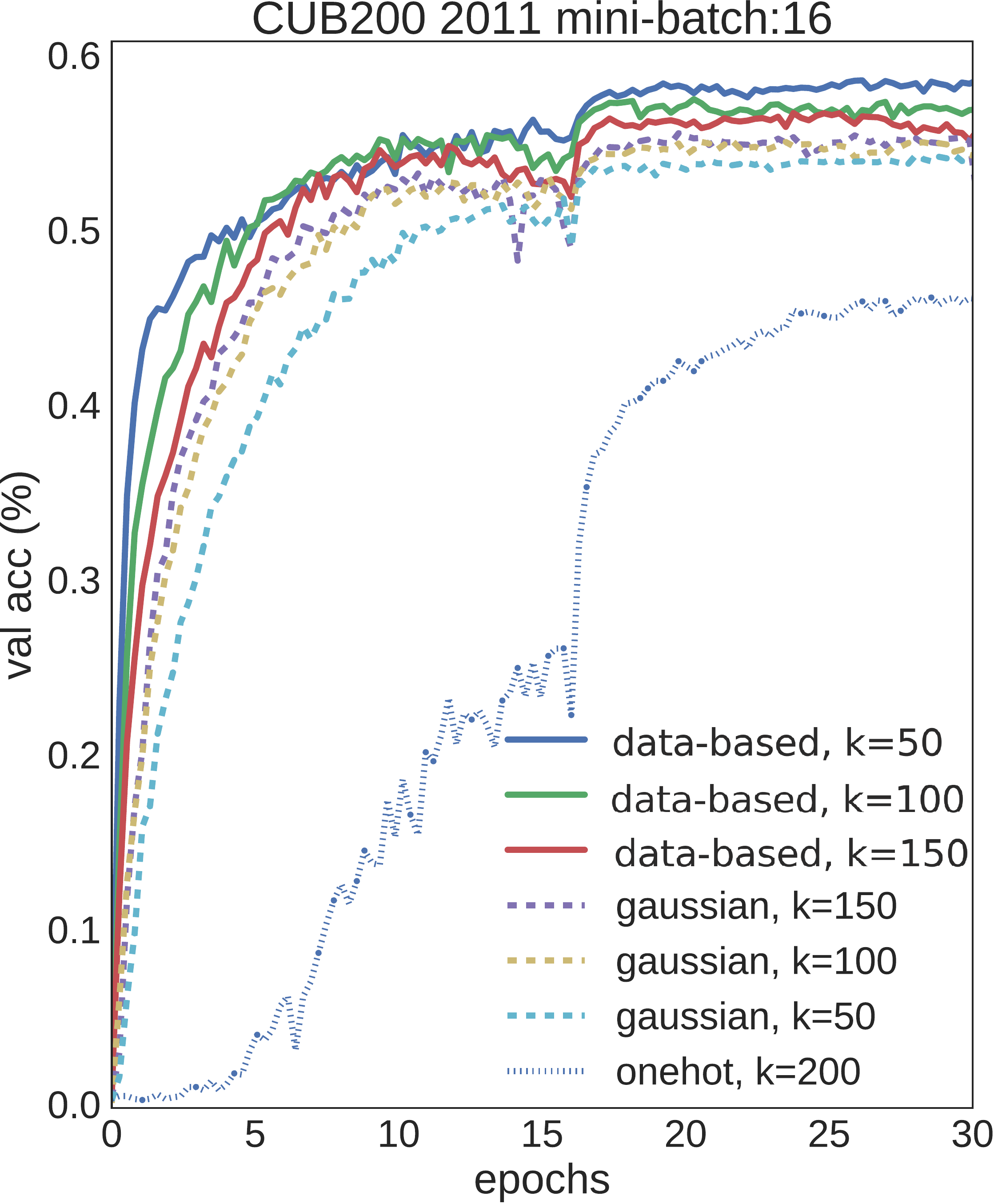}
\end{subfigure}
\begin{subfigure}[t]{0.45\textwidth}
\includegraphics[width=\textwidth, height=1\textwidth]{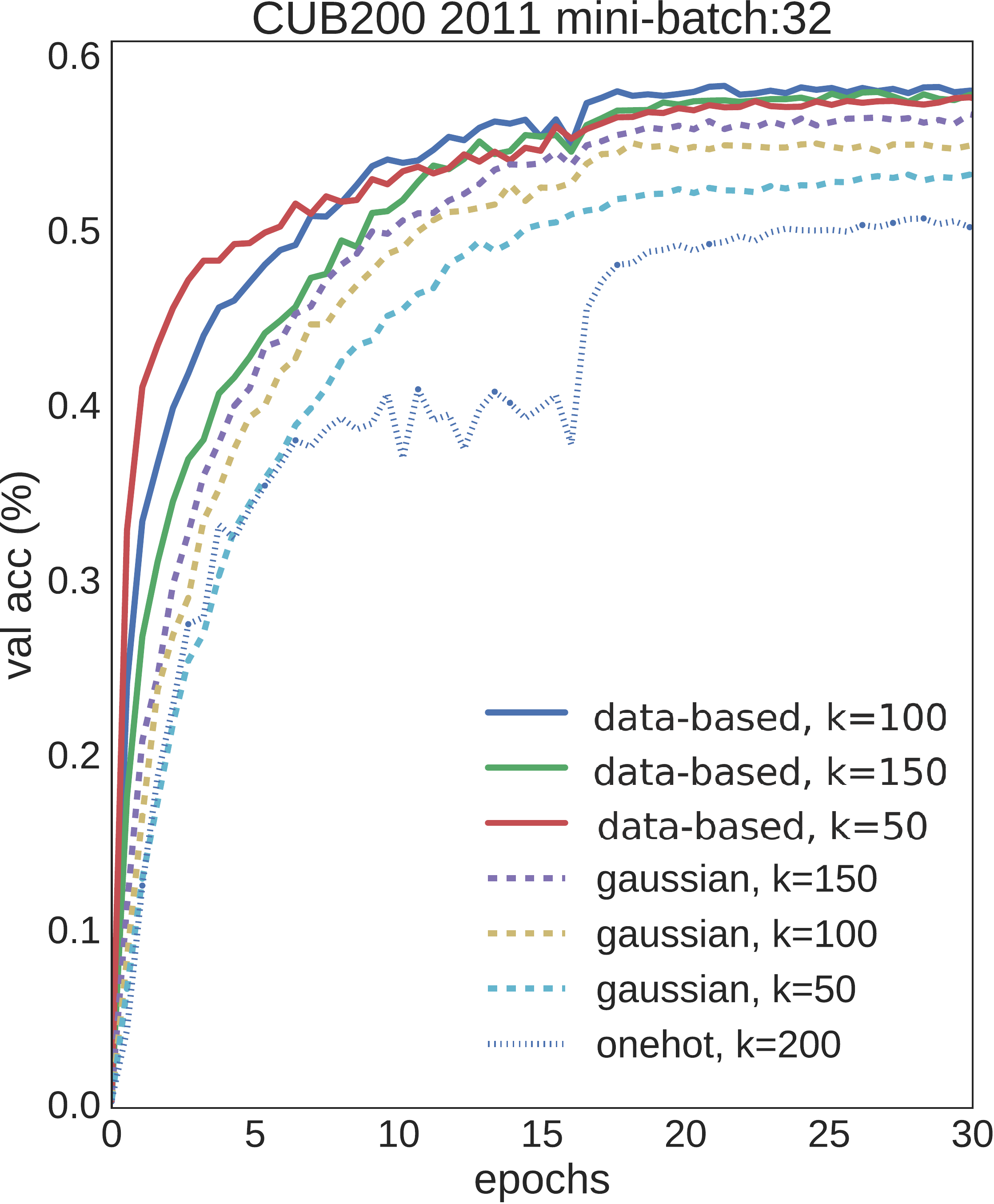}
\end{subfigure}
\begin{subfigure}[t]{0.45\textwidth}
\includegraphics[width=\textwidth, height=1\textwidth]{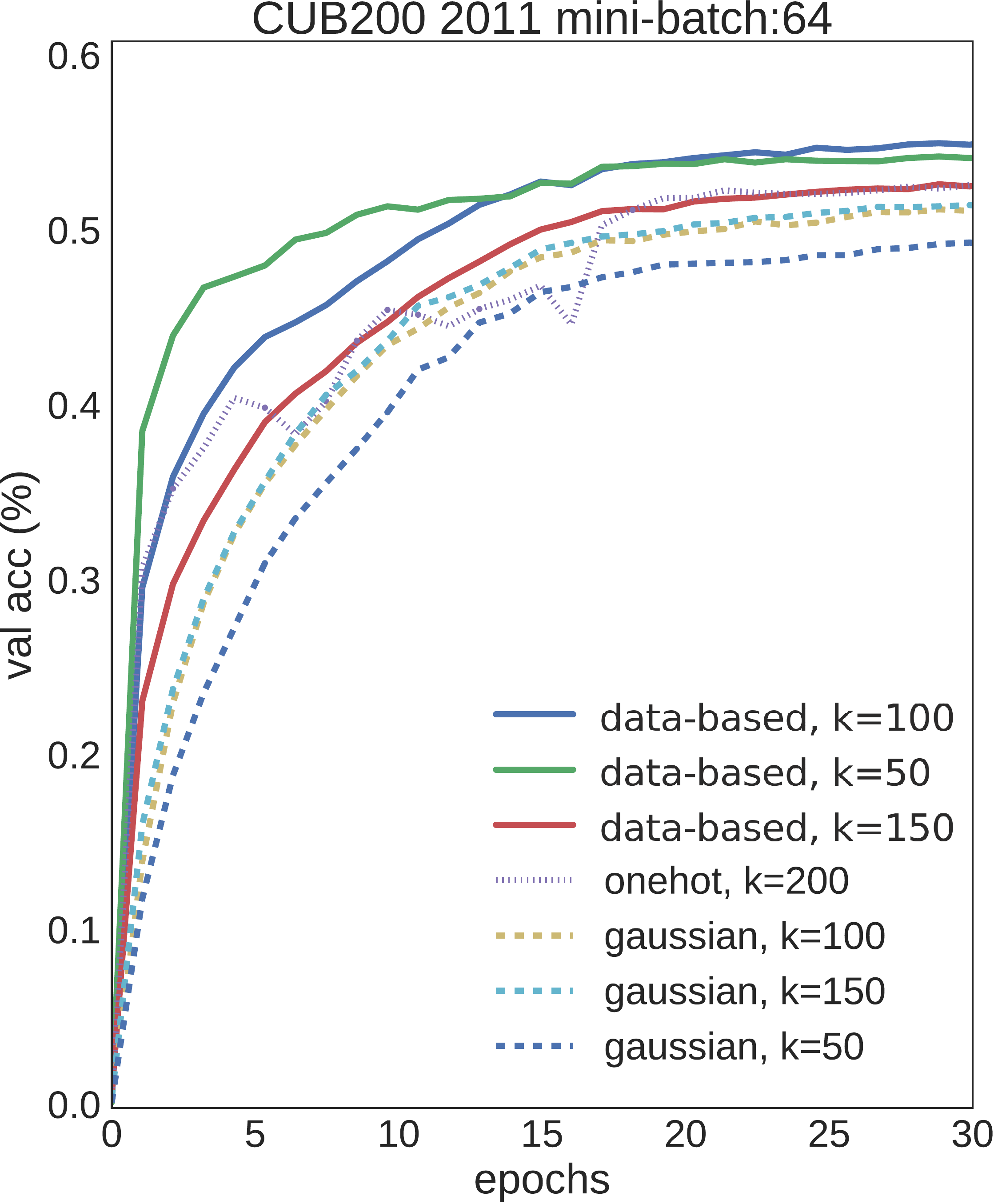}
\end{subfigure}
\begin{subfigure}[t]{0.45\textwidth}
\includegraphics[width=\textwidth, height=1\textwidth]{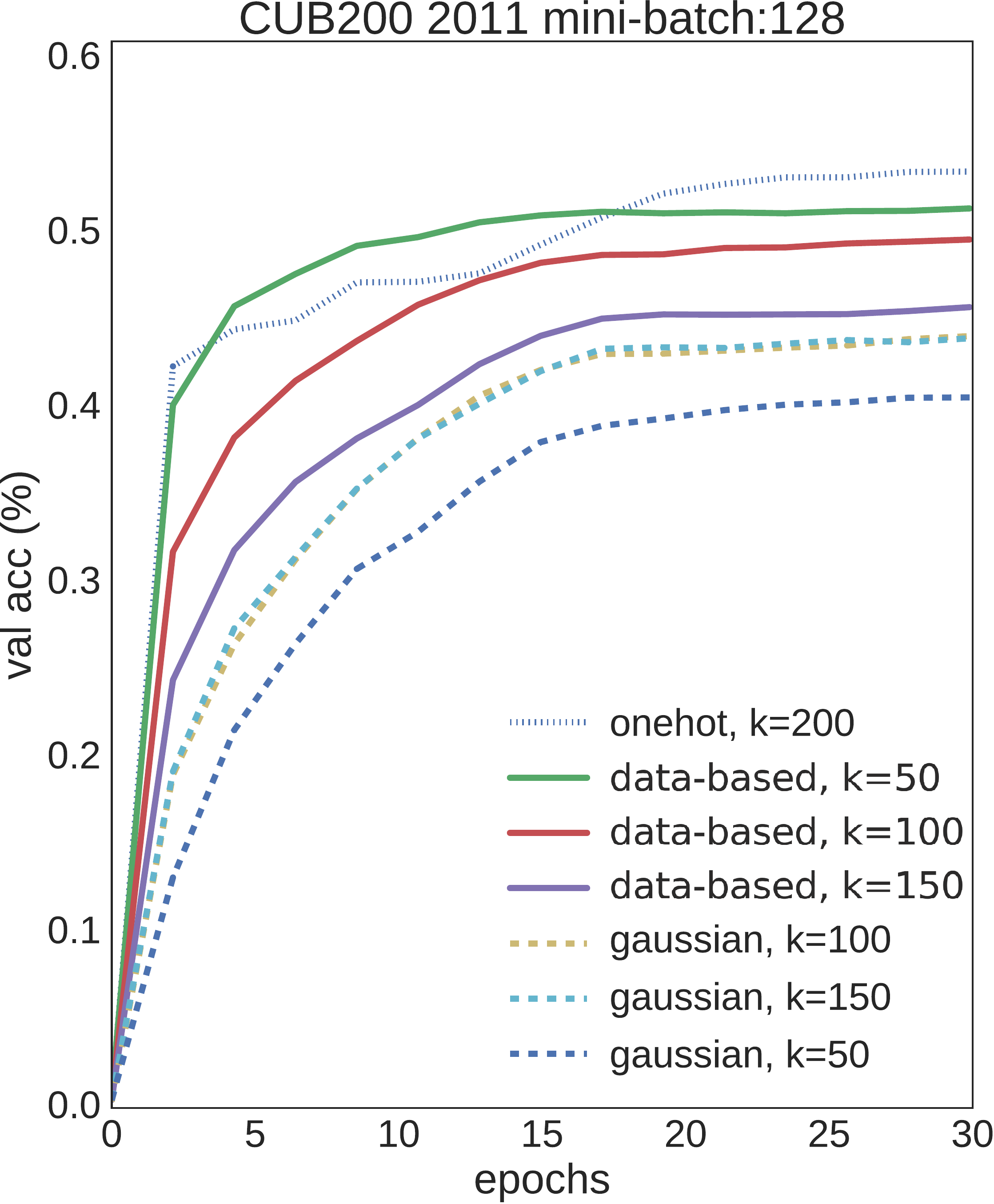}
\end{subfigure}
\caption{\textbf{Validation accuracy on CUB200.} Plots have been generated for different mini-batch sizes. (a) when mini-batch size is 16, the performance of one-hot encoding is dramatically reduced }
\label{fig:cub_accuracy_bs}\vspace{-0.4cm}
\end{figure*}

\noindent \textbf{CUB-200}. Figure \ref{fig:cub_accuracy_bs} shows that using small mini-batch sizes with data-based encodings largely outperforms the one-hot baseline for different code lengths when training a CNN on CUB-200 with data-dependent codes based on the raw eigenvalues of the class similarity matrix (best setting on CIFAR100). Moreover, in figure \ref{fig:cub_add_bits}, it can be seen that the data-based code matches the one-hot encoding with just the $25\%$ of the bits. As expected, the first bits correspond to the most discriminative partitions ordered by cut cost.
The class similarity matrix was built with the \texttt{fc7} outputs of a pre-trained network, but any other would also work if it reflects the inter-class relationships. \\

\begin{figure*}
\centering
\begin{subfigure}[t]{0.49\textwidth}
\includegraphics[width=\textwidth]{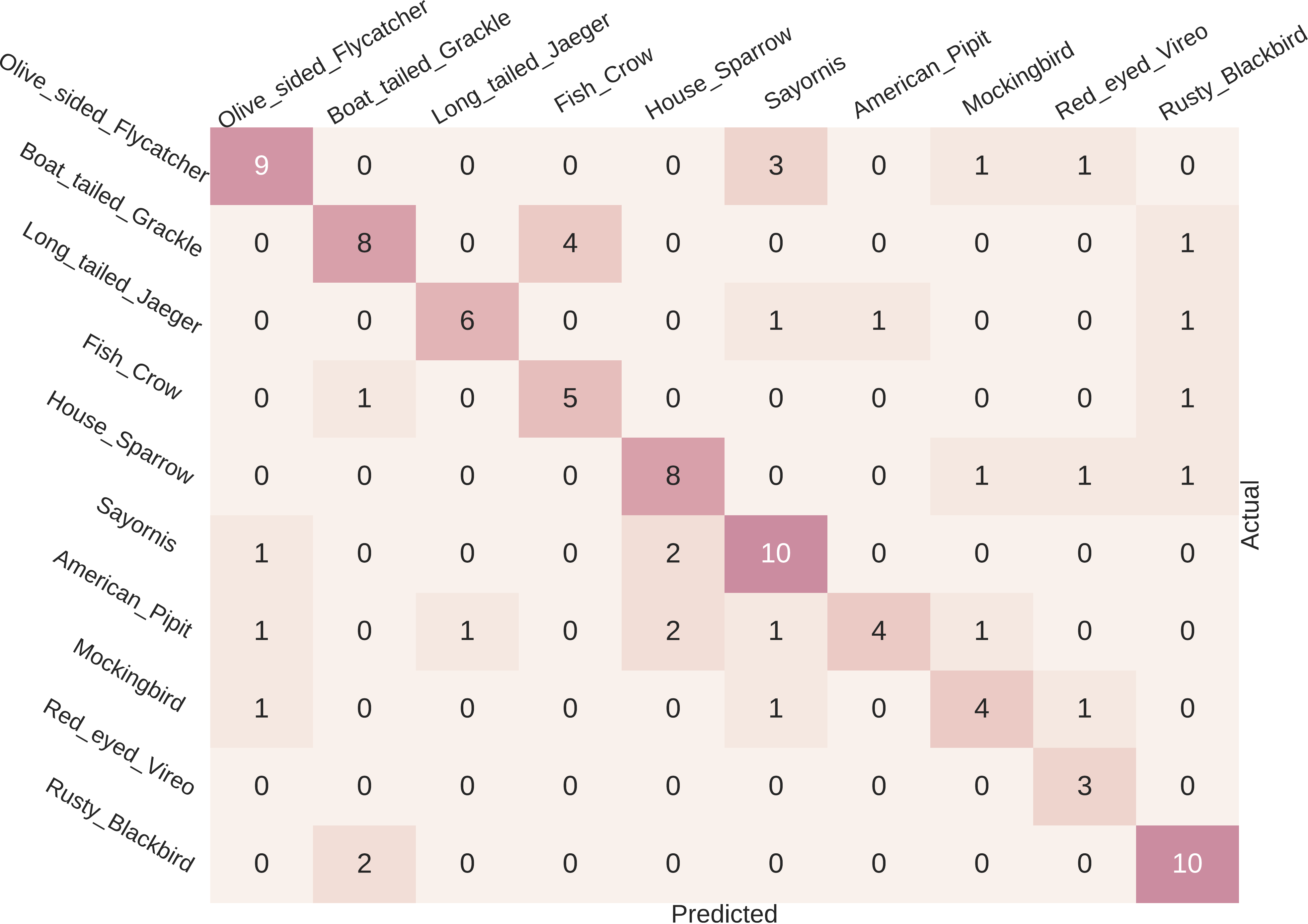}
\caption{One-hot encoding.}
\end{subfigure}
\begin{subfigure}[t]{0.49\textwidth}
\includegraphics[width=\textwidth]{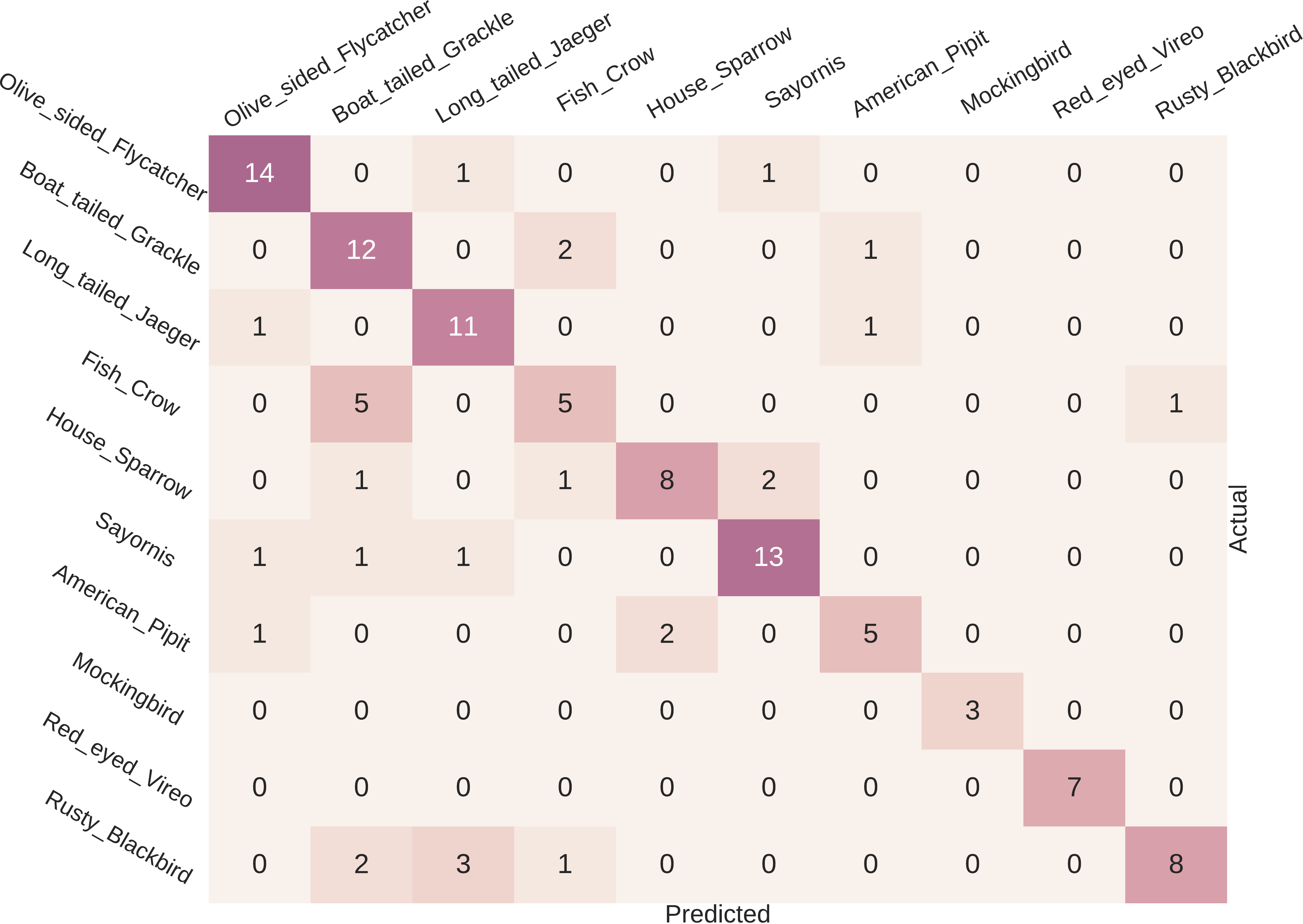}
\caption{Random encoding.}
\end{subfigure}
\begin{subfigure}[t]{0.49\textwidth}
\includegraphics[width=\textwidth]{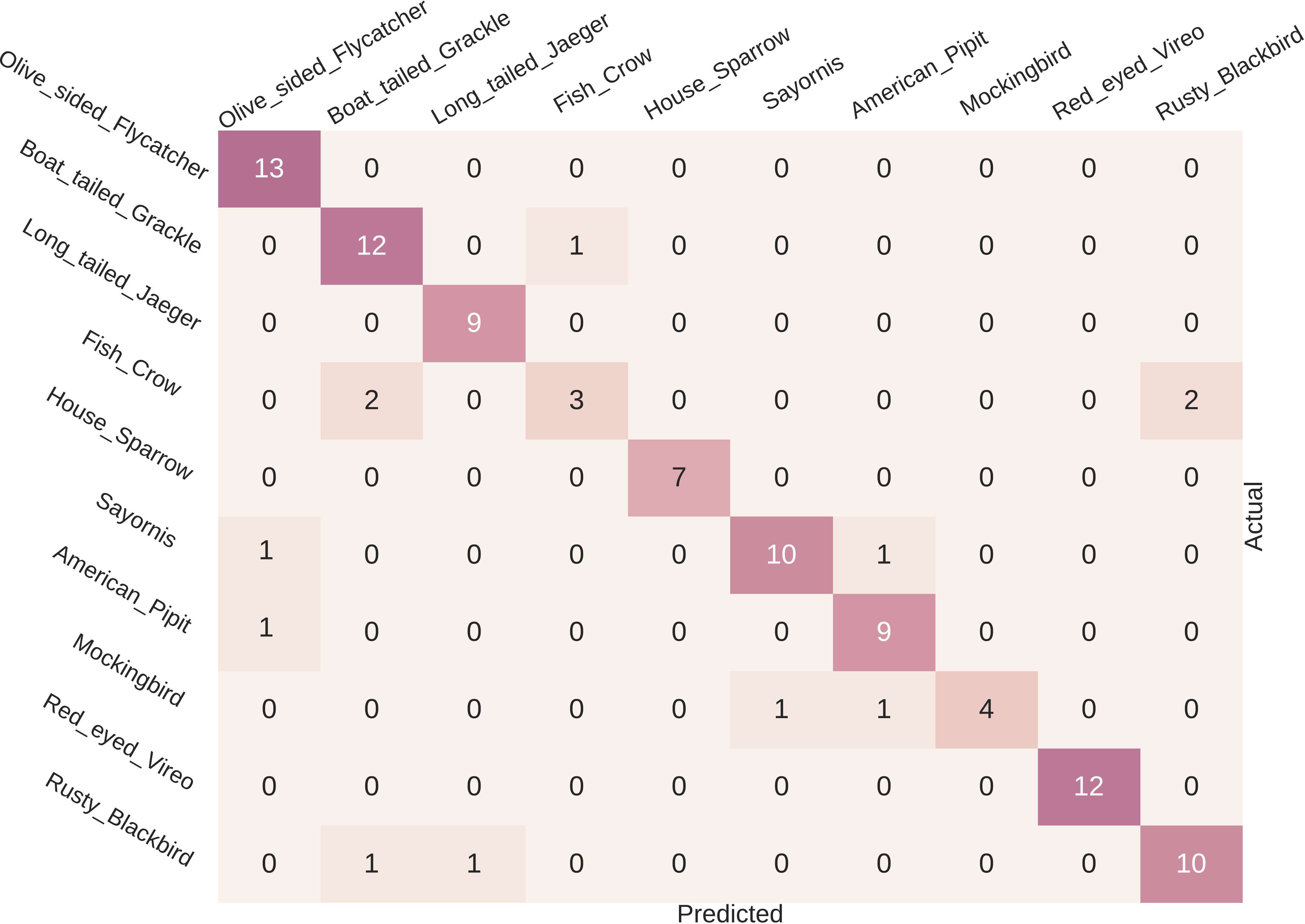}
\caption{Data-dependent code.}
\end{subfigure}
\caption{\textbf{Confusion matrices on CUB200-2011.} Alexnet trained with random codes sampled from a normal distribution (b) already advantage those trained with one-hot encoding (a) e.g., reducing the number of confusions of "Olive sided Flycatcher" with the rest of the classes. Moreover, data-dependent codes based on eigenrepresentations of the output space (d), can better discriminate even more classes, like "boat tailed Grackle" from "Fish crow". Samples for the classes in the confusion matrices are shown in figure \ref{fig:conf_classes}.}
\label{fig:cub_conf_mats}\vspace{-0.1cm}
\end{figure*}

\begin{figure}[!t]
\centering
\includegraphics[width=0.8\linewidth]{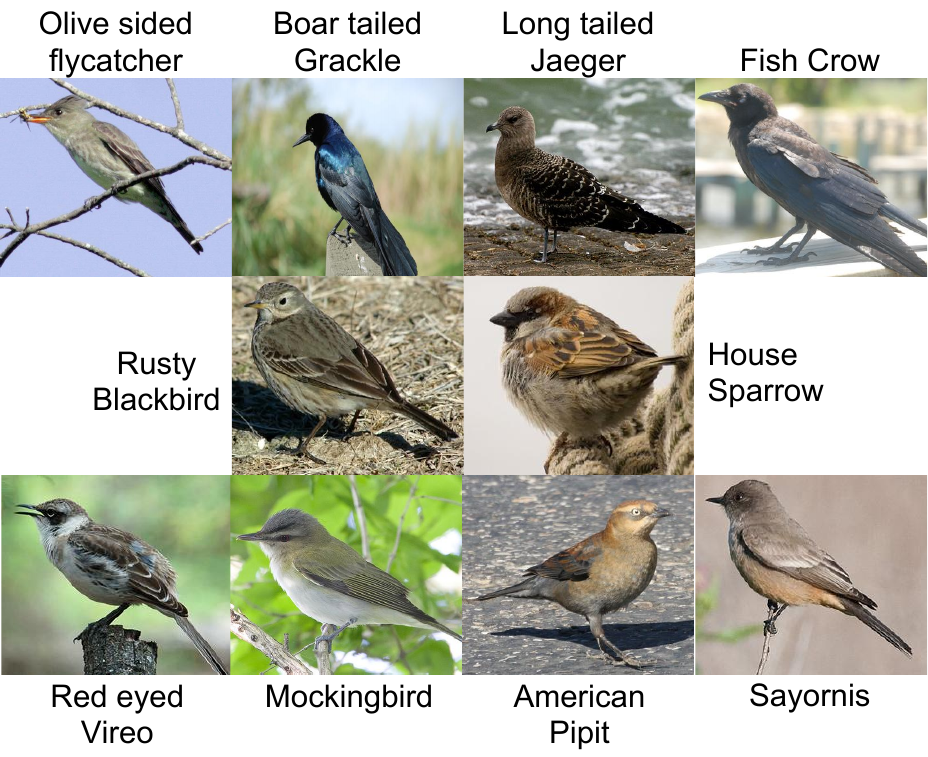}
\vspace{-0.1cm}
\caption{Confusion matrix classes.}
\label{fig:conf_classes}\vspace{-0.5cm}
\end{figure}

\begin{figure}[!t]
\centering
\includegraphics[width=0.75\textwidth]{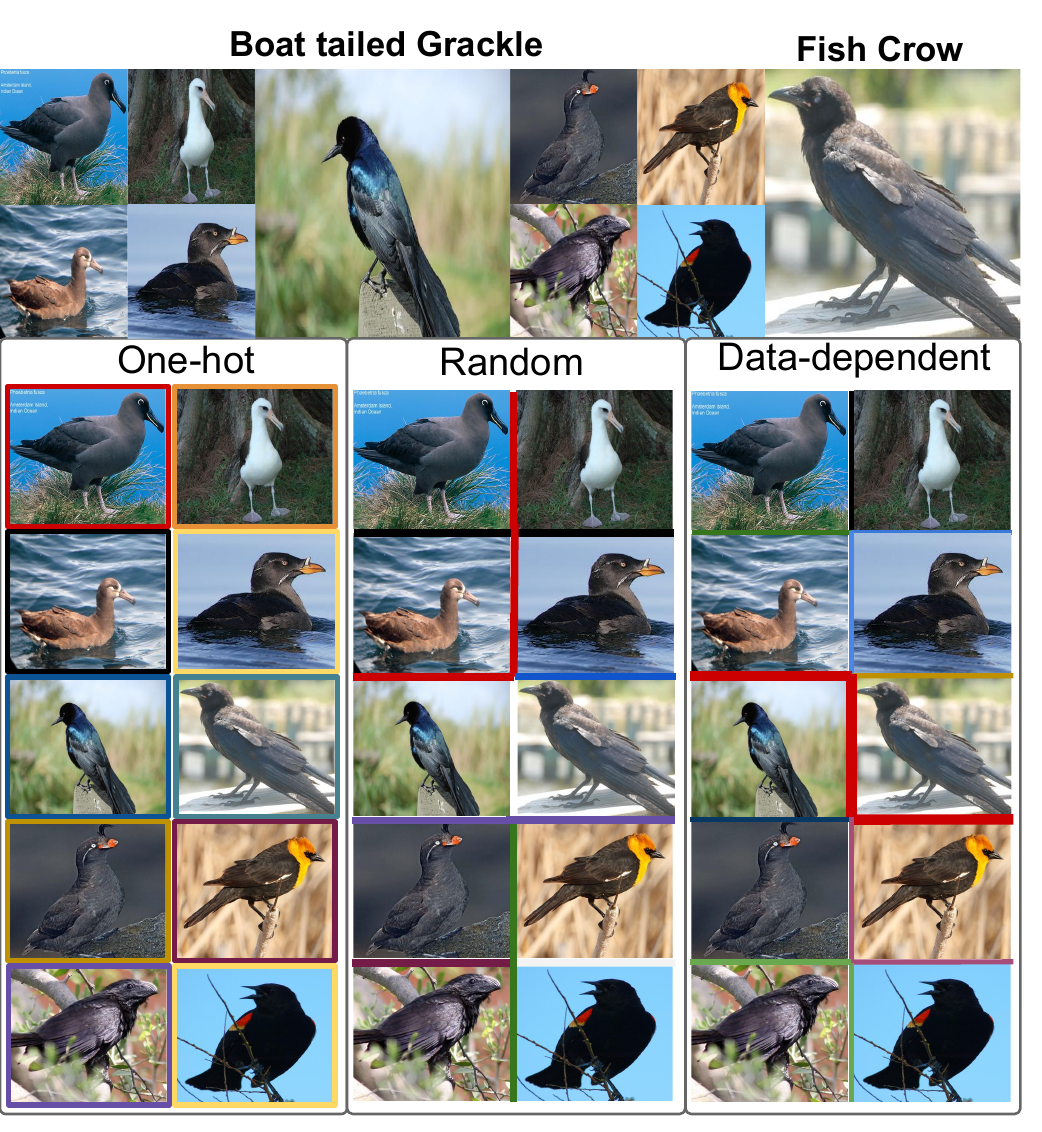}
\caption{\textbf{Classifying Boat tailed Grackle and Fish Crow.} One-hot encoding directly assigns labels to each of the examples. Random encoding partitions groups of classes into meta-classes systematically. Data-depending codes first group aquatic and non-aquatic birds, eliminating posterior confusions.}
\label{fig:spectral_advantages}
\end{figure}

Figure \ref{fig:cub_conf_mats} contains the confusion matrices for ten of the CUB-200 classes. Note that data-dependent encodings find low cost partitions, discriminating classes prone to be confused in the first stages of the hierarchy (the first encoding bits), and keeping those harder classification problems to the leafs. A comparison of one-hot, random and data-dependent encodings for the classification of "Fish crow" and "Grackle" is shown in figure \ref{fig:spectral_advantages}. 

We lastly verify the correspondence of the metaclasses found with data-dependent encodings by computing the Pearson Correlation Coefficient (CCP) between the columns of the code-matrix and the attributes associated to each of the CUB-200 classes, see table \ref{tab:attr_corr}. 

As expected, the data-dependent code finds a high-level partition that already discriminates both classes. One-hot, instead acts directly at the class level, without being explicitly based on shared attributes. On the other hand, random codes, although also based on metaclasses (attributes), do not guarantee that those metaclasses are the most discriminative ones.

\begin{table*}[!t]\scriptsize
\centering
\begin{subfigure}[!t]{\textwidth}
\centering
\begin{tabular}{|l|c|c|c|c|c|c|}
\hline
\textbf{Code bit} & \textbf{1} & \textbf{2} & \textbf{3} & \textbf{4} & \textbf{5} & \textbf{6} \\ \hline
\textbf{Attribute} & \begin{tabular}[c]{@{}c@{}}Belly-color\\ red\end{tabular} & \begin{tabular}[c]{@{}c@{}}Head-pattern\\ eyeline\end{tabular} & \begin{tabular}[c]{@{}c@{}}breast-color\\ blue\end{tabular} & \begin{tabular}[c]{@{}c@{}}bill-color\\ green\end{tabular} & \begin{tabular}[c]{@{}c@{}}head-pattern\\ unique\end{tabular} & \begin{tabular}[c]{@{}c@{}}crown-color\\ yellow\end{tabular} \\ \hline
\textbf{PCC} & 0.18 & 0.16 & 0.15 & 0.15 & 0.14 & 0.14 \\ \hline
\textbf{Attribute} & \begin{tabular}[c]{@{}c@{}}Tail-shape\\ rounded\end{tabular} & \begin{tabular}[c]{@{}c@{}}Under-tail-color\\ iridiscent\end{tabular} & \begin{tabular}[c]{@{}c@{}}biill-color\\ brown\end{tabular} & \begin{tabular}[c]{@{}c@{}}belly-color\\ pink\end{tabular} & \begin{tabular}[c]{@{}c@{}}bill-shape\\ all-purpose\end{tabular} & \begin{tabular}[c]{@{}c@{}}tail-shape\\ forked\end{tabular} \\ \hline
\textbf{PCC} & -0.22 & -0.17 & -0.17 & -0.16 & -0.18 & -0.18 \\ \hline
\end{tabular}
\subcaption{Random Code.}
\end{subfigure}
\begin{subfigure}[t!]{\textwidth}
\centering
\begin{tabular}{|l|c|c|c|c|c|c|}
\hline
\textbf{Code bit} & \textbf{1} & \textbf{2} & \textbf{3} & \textbf{4} & \textbf{5} & \textbf{6} \\ \hline
\textbf{Attribute} & \begin{tabular}[c]{@{}c@{}}shape\\ perching-like\end{tabular} & \begin{tabular}[c]{@{}c@{}}primary-color\\ yellow\end{tabular} & \begin{tabular}[c]{@{}c@{}}back-color\\ black\end{tabular} & \begin{tabular}[c]{@{}c@{}}bill-color\\ black\end{tabular} & \begin{tabular}[c]{@{}c@{}}throat-color\\ yellow\end{tabular} & \begin{tabular}[c]{@{}c@{}}upperpart-color\\ white\end{tabular} \\ \hline
\textbf{PCC} & 0.79 & 0.64 & 0.50 & 0.44 & 0.53 & 0.55 \\ \hline
\textbf{Attribute} & size:medium & \begin{tabular}[c]{@{}c@{}}upper-tail-color\\ brown\end{tabular} & \begin{tabular}[c]{@{}c@{}}wing-color\\ grey\end{tabular} & \begin{tabular}[c]{@{}c@{}}primary-color\\ red\end{tabular} & \begin{tabular}[c]{@{}c@{}}primary-color\\ rufous\end{tabular} & \begin{tabular}[c]{@{}c@{}}belly-color\\ black\end{tabular} \\ \hline
\textbf{PCC} & -0.73 & -0.56 & -0.58 & -0.38 & -0.42 & -0.48 \\ \hline
\end{tabular}
\subcaption{Data-dependent code.}
\end{subfigure}
\caption{\textbf{Top CUB200 attributes by correlation with the code.} Random codes do not show relevant correlations with the data attributes, while data-dependent codes are visibly correlated with the attributes. Concretely, the first bit of the code, i.e. the partition with the lowest cut cost, is highly correlated with shape and size attributes ($0.79$). The sign of the PPC indicates the expected side of the bi-partition associated for the attribute. As expected, the PPC coefficient decreases in absolute value as the cut cost increases, since higher bits correspond to increasingly difficult partitions.}
\label{tab:attr_corr}\vspace{-0.4cm}
\end{table*}

\section{Conclusion}\label{sec:conclusion}
In this work, output codes are integrated with the training of deep CNNs on large-scale datasets. We found that CNNs trained on CIFAR-100, CUB200, Imagenet, and MIT Places using our approach show less sparsity at the output neurons. As a result, models trained with our approach showed more robust gradient estimates and faster convergence rates than those trained with the prevalent one-hot encoding at a small cost, especially for huge label spaces. As a side effect, CNNs trained with our approach can use smaller minibatch sizes, lowering the memory consumption. Moreover, we showed that training with data-dependent codes based on eigenrepresentations of the class space allows for more efficient, hierarchical representations, achieving lower error rates than those trained with data-independent output codes.

\section*{Acknowledgements}
Authors acknowledge the support of the Spanish project TIN2015-65464-R (MINECO FEDER), the 2016FI\_B 01163 grant (Secretaria d’Universitats i Recerca del Departament d’Economia i Coneixement de la Generalitat de Catalunya), and the COST Action IC1307 iV\&L Net (European Network on Integrating Vision and Language) supported by COST (European Cooperation in Science and Technology). We also gratefully acknowledge the support of NVIDIA Corporation with the donation of a Tesla K40 GPU and a GTX TITAN GPU, used for this research.

\bibliographystyle{abbrvnat}
\bibliography{egbib}

\end{document}